\documentclass[journal]{IEEEtran}

\usepackage{cite}
\usepackage{amsmath,amssymb,amsfonts}
\usepackage{graphicx}
\usepackage{textcomp}
\usepackage{xcolor}
\usepackage{booktabs}
\usepackage{multirow}
\usepackage{diagbox}
\usepackage{hyperref}
\usepackage{array}
\usepackage{url}
\usepackage{bm}
\usepackage{balance}
\usepackage{algorithm}
\usepackage{algorithmic}
\usepackage{subcaption}
\usepackage{float}

\usepackage{hyperref}
\hypersetup{
    colorlinks=true,
    linkcolor=blue,
    citecolor=blue,
    urlcolor=blue,
}

\begin{document}
\bstctlcite{IEEEexample:BSTcontrol}

\title{Sparse Gain Radio Map Reconstruction With Geometry Priors and Uncertainty-Guided Measurement Selection}

\author{Zhihan Zeng, \IEEEmembership{Graduate Student Member, IEEE}, Ning Wei, \IEEEmembership{Member, IEEE},\\
Kaihe Wang, \IEEEmembership{Graduate Student Member, IEEE}, Phee Lep Yeoh, \IEEEmembership{Senior Member, IEEE},\\ Fei Xu,
Yue Xiu, \IEEEmembership{Member, IEEE}, Zhongpei Zhang \\

\thanks{Zhihan Zeng, Ning Wei, Yue Xiu and Zhongpei Zhang are with the National Key Laboratory of Wireless Communications, University of Electronic Science and Technology of China (UESTC), Chengdu 611731, China (E-mail: 202511220608@std.uestc.edu.cn, wn@uestc.edu.cn, xiuyue12345678@163.com, zhangzp@uestc.edu.cn). Kaihe Wang is with the University of Electronic Science and Technology of China (UESTC), Chengdu 611731, China (E-mail: khewang@yeah.net). Phee Lep Yeoh is with the School of Science, Technology and Engineering, University of the Sunshine Coast, Moreton Bay Campus, Petrie, QLD 4502, Australia (E-mail:pyeoh@usc.edu.au). Fei Xu is with the ZGC Institute of Ubiquitous-X Innovation and Applications, Beijing, China (E-mail: xufei@zgc-xnet.com). The corresponding author is Ning Wei.}
}

\maketitle
\begin{abstract}
Radio maps are important for environment-aware wireless communication, network planning, and radio resource optimization. However, dense radio map construction remains challenging when only a limited number of measurements are available, especially in complex urban environments with strong blockages, irregular geometry, and restricted sensing accessibility. Existing methods have explored interpolation, low-rank cartography, deep completion, and channel knowledge map (CKM) construction, but many of these methods insufficiently exploit explicit geometric priors or overlook the value of predictive uncertainty for subsequent sensing.
In this paper, we study sparse gain radio map reconstruction from a geometry-aware and active sensing perspective. We first construct \textbf{UrbanRT-RM}, a controllable ray-tracing benchmark with diverse urban layouts, multiple base-station deployments, and multiple sparse sampling modes. We then propose \textbf{GeoUQ-GFNet}, a lightweight network that jointly predicts a dense gain radio map and a spatial uncertainty map from sparse measurements and structured scene priors. The predicted uncertainty is further used to guide active measurement selection under limited sensing budgets.
Extensive experiments show that our proposed GeoUQ-GFNet method achieves strong and consistent reconstruction performance across different scenes and transmitter placements generated using UrbanRT-RM. Moreover, uncertainty-guided querying provides more effective reconstruction improvement than non-adaptive sampling under the same additional measurement budget. These results demonstrate the effectiveness of combining geometry-aware learning, uncertainty estimation, and benchmark-driven evaluation for sparse radio map reconstruction in complex urban environments.
\end{abstract}

\section{Introduction}
\IEEEPARstart{T}{he} evolution toward sixth-generation (6G) wireless networks is expected to fundamentally transform communication systems from connectivity-centric infrastructures into deeply integrated intelligent platforms that jointly support communication, sensing, computing, control, and autonomous decision making \cite{11353414,11355857,11316665,Saad2020Vision6G,Nguyen2022Survey6GIoT}. In this vision, future networks are not only required to deliver higher spectral efficiency, lower latency, and broader coverage, but also to acquire, represent, and exploit environmental knowledge in a native and scalable manner. This trend is consistent with the IMT-2030 vision, where intelligence, ubiquitous connectivity, and environment-aware operation are regarded as key characteristics of beyond-5G and 6G systems.\cite{Sun2026KnowledgeDrivenDL6G} Against this background, spatial channel knowledge representation is becoming an essential enabler for predictive and proactive wireless operation rather than a secondary support function.

Radio maps provide a spatial representation of wireless propagation conditions over a geographic region and have become increasingly important for future intelligent wireless systems\cite{11098592,11207524,Liu2025CKM6GSurvey,Wang2025CKMAidedChannelPrediction,Yang2025RadioMapBeamformingReducedPilots,Liu2025CKMAssistedRoutingLAE}. By describing channel-related quantities such as path loss, received power, or channel gain as functions of location, radio maps can support a wide range of tasks, including coverage analysis, deployment optimization, beam management, pilot reduction, routing, and environment-aware sensing \cite{Romero2024AerialBaseStationPlacementRadioMaps,Wu2024EnvironmentAwareHybridBeamformingCKM,Shi2025CKMBeamTrainingUncertainty,Sun2026KnowledgeDrivenDL6G,Nguyen2022Survey6GIoT,Kato2019AISAGIN,Liu2018SAGINSurvey,Saad2020Vision6G}. As wireless networks evolve toward more adaptive, data-driven, and environment-aware operation, radio maps are shifting from an auxiliary planning tool to a core infrastructure component for perception-assisted communication \cite{Cai2025DeepSpaceChannelEstimation,Gao2025LEODeepSpacePerformance,zeng2026gackan,zeng2026phygmoe,zeng2026skanet,zeng2026jsrgfnet}.

Despite their promise, accurate radio map construction remains difficult in realistic environments. In principle, a dense map requires exhaustive measurements over the entire target area, but such acquisition is often prohibitively expensive in practice due to limited sensing budgets, constrained accessibility, and dynamic deployment requirements. This difficulty is especially pronounced in urban environments, where propagation is jointly shaped by blockage, building geometry, line-of-sight conditions, and transmitter placement. As a result, the practical problem is not dense measurement-based mapping, but sparse-to-dense reconstruction, namely how to infer a reliable dense radio map from only a small subset of accessible observations \cite{Wu2018AutomaticRadioMapAdaptation}.

A long line of research has focused on radio map estimation and channel gain cartography using interpolation, tomography, kernel regression, and structure-exploiting recovery. Representative examples include radio tomography and cartography methods based on spatial inversion, low-rank structure, sparsity priors, and side information \cite{Martin2014AccuracyVsResolutionRTI,Romero2018BlindRadioTomography,Lee2017ChannelGainCartographyLowRankSparse,Lee2019AdaptiveBayesianRadioTomography,Kasparick2016KernelOnlineCoverageMap,Romero2017PSDMapQuantizedMeasurements,Teganya2019LocationFreeSpectrumCartography,Sato2021SpaceFrequencyRadioMap,Romero2024TheoreticalAnalysisRME,Bazerque2010DistributedSpectrumSensingSparsity}. These methods provide useful modeling insights, but they usually rely on relatively rigid assumptions about spatial smoothness, kernel correlation, or linear structure, which limits their effectiveness in highly heterogeneous urban scenes.

More recently, deep learning has substantially advanced radio map reconstruction by learning complex nonlinear propagation patterns directly from data. Early deep models such as RadioUNet and deep completion autoencoders showed that convolutional networks can effectively recover dense maps from sparse spatial observations and environmental inputs \cite{11346858,11316633,11108293,10797657,11220909,Levie2021RadioUNet,Teganya2022DeepCompletionAutoencoders,Shrestha2022DeepSpectrumCartography,Roger2024DLRadioMapV2X,Jaiswal2026TransferLearningIndoorRadioMap,Liao2026RadioKAN,Yang2025AIGCRadioMapLAE,Li2024RadioMapPredictiveUAV,Mu2021IRSRobotPathPlanningRadioMap, Jaiswal2026TLMoERadioMap}. Subsequent studies further explored graph-based modeling, inpainting, deformable attention, restricted-area reconstruction, and generative diffusion-based radio map construction \cite{Li2024RadioGAT,Zhang2024RadiomapInpaintingRestrictedAreas,Liu2026DATUnetRadioMap,Wang2025RadioDiff,Chen2025Urban3DRadioMapSparse,Zhao2025RMCDDRUAV}. In parallel, the emerging channel knowledge map (CKM) literature has expanded the construction paradigm toward image inpainting, super-resolution, diffusion generation, flow matching, beam-aware mapping, and cross-access-point inference \cite{Jin2025I2IInpaintingCKM,Wang2025SRResNetCKM,Fu2025PartialObservationDiffusionCKM,Huang2025GuidedFlowMatchingCKM,Zhu2025PhysicsInspiredDiffusionCKM,Wang2025BeamCKM,Wang2025CodebookAwareCKM,Zhao2026BeamCKMDiff,Dai2025CrossAPCKM,Xu2024HowMuchDataCKM}. These advances clearly demonstrate the potential of data-driven radio environment reconstruction.

However, two important gaps remain. First, although many recent methods achieve strong reconstruction accuracy, explicit integration of fine-grained geometric priors is still insufficient in sparse urban mapping. Existing works have started to exploit environmental information from depth maps, point clouds, and higher-dimensional scene representations \cite{Zhang2024RadiomapInpaintingRestrictedAreas,Wang2025PointCloudCKM,Wang2025TowardsPreciseCKM,Zhou2025BiWGS6DCKM}, yet for sparse gain map completion under limited observations, the question of how to effectively fuse obstacle structure, building height, relative transmitter geometry, accessibility constraints, and sparse measurements within a unified lightweight architecture is still underexplored. Second, most existing reconstruction methods mainly optimize one-shot prediction quality, whereas in practical measurement campaigns the more important question is often where to measure next. In this sense, predictive uncertainty is not merely an auxiliary output but a decision variable for active sensing. Although Bayesian active learning and adaptive tomography have demonstrated the value of uncertainty-aware sampling in radio map reconstruction \cite{Polyzos2024BayesianActiveLearningRadioMap,Lee2019AdaptiveBayesianRadioTomography,Shrestha2023SpectrumSurveying}, the integration of geometry-aware deep reconstruction and uncertainty-guided measurement selection remains limited.

Motivated by these observations, this paper studies sparse gain radio map reconstruction from a problem-oriented perspective centered on geometry priors and uncertainty-guided sensing. We first build UrbanRT-RM, a controllable ray-tracing benchmark over diverse synthetic urban scenes, enabling systematic analysis across different scene categories, base-station deployments, and sparse sampling modes. We then propose GeoUQ-GFNet, a geometry-aware and uncertainty-aware reconstruction network that jointly predicts a dense gain radio map and a spatial uncertainty map from sparse observations and structured scene priors. The model introduces a Geometry-Gated Front-End to adaptively fuse heterogeneous inputs, a lightweight multi-stage encoder combining Ghost modules \cite{Han2020GhostNet} and Grid-KAN-style nonlinear enhancement inspired by Kolmogorov-Arnold network (KAN) representations \cite{Liu2025KAN}, and dual output heads for gain prediction and uncertainty estimation. The predicted uncertainty is further used to guide active measurement selection under limited sensing budgets.

The main contributions of this paper are summarized as follows.
\begin{itemize}
    \item We introduce a geometry-aware and uncertainty-aware formulation of sparse gain radio map reconstruction that explicitly accounts for accessibility-constrained sensing and connects reconstruction with sequential measurement decision.
    
    \item We develop UrbanRT-RM, a controllable ray-tracing benchmark that disentangles the effects of scene geometry, transmitter deployment, and sparse sampling pattern on reconstruction difficulty.
    
    \item We propose GeoUQ-GFNet, a lightweight geometry-aware network with a geometry-gated front-end and dual gain--uncertainty heads for joint dense radio map reconstruction and uncertainty estimation from sparse observations.
    
    \item We further develop an uncertainty-guided measurement selection strategy and show that the learned uncertainty improves sample efficiency over non-adaptive querying under the same sensing budget.
\end{itemize}

The rest of this paper is organized as follows. Section~\ref{sec_related} reviews related work on radio map reconstruction, CKM construction, and uncertainty-aware active sensing. Section~\ref{sec_model} introduces the system model and problem formulation. Section~\ref{sec_method} presents the proposed GeoUQ-GFNet architecture and the uncertainty-guided measurement selection strategy. Section~\ref{sec_experiments} describes the UrbanRT-RM dataset generation process and the experimental setup. Section~\ref{sec_results} reports the reconstruction and active sensing results. Finally, Section~\ref{sec_conclusion} concludes the paper.

\section{Related Work}
\label{sec_related}

\subsection{Traditional Radio Map Reconstruction and Cartography}
Early studies on radio map estimation mainly relied on interpolation, tomography, and structure-based recovery. Classical radio tomography works analyzed the relationship between measurement geometry, reconstruction accuracy, and spatial resolution, and later extended the framework toward blind estimation and adaptive Bayesian sensing \cite{Martin2014AccuracyVsResolutionRTI,Wilson2011VarianceBasedRTI,Romero2018BlindRadioTomography,Lee2019AdaptiveBayesianRadioTomography}. In parallel, channel gain cartography studied the recovery of large-scale spatial gain fields by exploiting structural priors such as low rank and sparsity \cite{Lee2017ChannelGainCartographyLowRankSparse}. Related studies further investigated kernel-based online reconstruction, quantized measurement learning, location-free cartography, and space-frequency interpolation \cite{Kasparick2016KernelOnlineCoverageMap,Romero2017PSDMapQuantizedMeasurements,Teganya2019LocationFreeSpectrumCartography,Sato2021SpaceFrequencyRadioMap}. These methods established an important theoretical and methodological foundation for radio map reconstruction. However, they usually rely on explicit statistical assumptions or handcrafted spatial models, which makes it difficult to capture the highly nonlinear propagation effects caused by complex urban geometry, irregular blockage, and heterogeneous sensing support.

\subsection{Deep Learning and Generative Construction of Radio Maps and CKMs}
Deep learning has significantly improved sparse radio map reconstruction by learning nonlinear mappings from environment information and sparse observations to dense propagation maps. Representative radio map estimation methods include RadioUNet, deep completion autoencoders, tensor completion with learned neural models, graph-attention-based reconstruction, deformable-attention modeling, and several task-specific variants for V2X, UAV, and dynamic radio environments \cite{Levie2021RadioUNet,Teganya2022DeepCompletionAutoencoders,Shrestha2022DeepSpectrumCartography,Roger2024DLRadioMapV2X,Li2024RadioGAT,Liu2026DATUnetRadioMap,Zhao2025RMCDDRUAV,Chen2025DynamicSpectrumCartography,Wang2025RadioDiff}. These studies show that learned models can substantially outperform classical recovery techniques when sufficient data and environmental context are available.

In the emerging CKM literature, recent methods have reformulated the task as image completion, super-resolution, or generative modeling. For example, image-to-image inpainting has been used to recover dense channel knowledge maps from sparse observations \cite{Jin2025I2IInpaintingCKM}, while super-resolution style reconstruction has been explored to enhance map resolution from sparse or coarse inputs \cite{Wang2025SRResNetCKM}. Generative approaches based on diffusion models and flow matching further extend the construction paradigm beyond deterministic regression, enabling more flexible distribution modeling for partially observed or even observation-free scenarios \cite{Fu2025PartialObservationDiffusionCKM,Huang2025GuidedFlowMatchingCKM,Zhu2025PhysicsInspiredDiffusionCKM}. In addition, CKM construction has been broadened from single-antenna gain maps to multi-antenna and beam-aware settings, where beamforming vectors, codebooks, and cross-AP information are explicitly incorporated into the mapping process \cite{Wang2025BeamCKM,Wang2025CodebookAwareCKM,Zhao2026BeamCKMDiff,Dai2025CrossAPCKM}. Recent works also explored point-cloud-based CKM construction, 2D-to-3D environmental representation, and 6D scene-aware modeling \cite{Wang2025PointCloudCKM,Wang2025TowardsPreciseCKM,Zhou2025BiWGS6DCKM}. Public benchmarks such as CKMImageNet and SpectrumNet further promote standardized evaluation for AI-based radio environment learning \cite{Wu2025CKMImageNet,Zhang2025SpectrumNet}.

Although these studies clearly demonstrate the importance of environment-aware modeling, many existing methods are either relatively heavy, tailored to beam-aware or generative settings, or primarily optimized for one-shot reconstruction. For the sparse gain radio map task considered in this paper, there is still a need for a lightweight architecture that can explicitly and effectively fuse structured geometry priors with sparse measurements under accessibility constraints.

\subsection{Uncertainty-Aware Reconstruction and Active Measurement Selection}
In practical radio map acquisition, the challenge is not only to reconstruct a dense map from limited data, but also to decide how to allocate future measurements efficiently. This has motivated active sensing and uncertainty-aware sampling methods. Bayesian active learning for radio map reconstruction has shown that uncertainty estimates can substantially improve sample efficiency by prioritizing informative sensing locations \cite{Polyzos2024BayesianActiveLearningRadioMap}. Related ideas also appear in adaptive Bayesian radio tomography and active UAV-based spectrum surveying, where uncertainty or posterior information is used to guide sequential measurements \cite{Lee2019AdaptiveBayesianRadioTomography,Shrestha2023SpectrumSurveying,Zeng2021SNARMUAV}. These works demonstrate that uncertainty is highly valuable for measurement planning. However, most existing uncertainty-aware methods are built on Gaussian process or Bayesian inference frameworks and are not tightly integrated with modern geometry-aware deep reconstruction networks. As a result, the interaction between structured scene priors, learned uncertainty, and active measurement selection remains insufficiently studied in sparse urban radio map reconstruction.

\subsection{Position of This Work}
Compared with the above studies, this work differs in three main aspects. First, we focus on \emph{sparse gain radio map reconstruction} in complex urban environments with explicit accessibility constraints, which is closely aligned with realistic outdoor measurement campaigns. Second, instead of using geometry merely as an auxiliary input, we design a dedicated geometry-gated fusion mechanism to modulate sparse observation features using structural priors and sensing masks. Third, we explicitly connect \emph{uncertainty estimation} with \emph{active measurement selection}, so that the model output is useful not only for one-shot reconstruction but also for subsequent sensing decisions under limited sensing budgets. In addition, we support the study with UrbanRT-RM, a controllable ray-tracing benchmark designed for scene-level and deployment-level evaluation of sparse radio map reconstruction. In this sense, the proposed framework sits at the intersection of geometry-aware deep radio map learning, benchmark-driven evaluation, and uncertainty-aware active sensing, which remains insufficiently explored in the current literature.

\begin{figure*}[t!]
    \centering
    \includegraphics[width=0.85\linewidth]{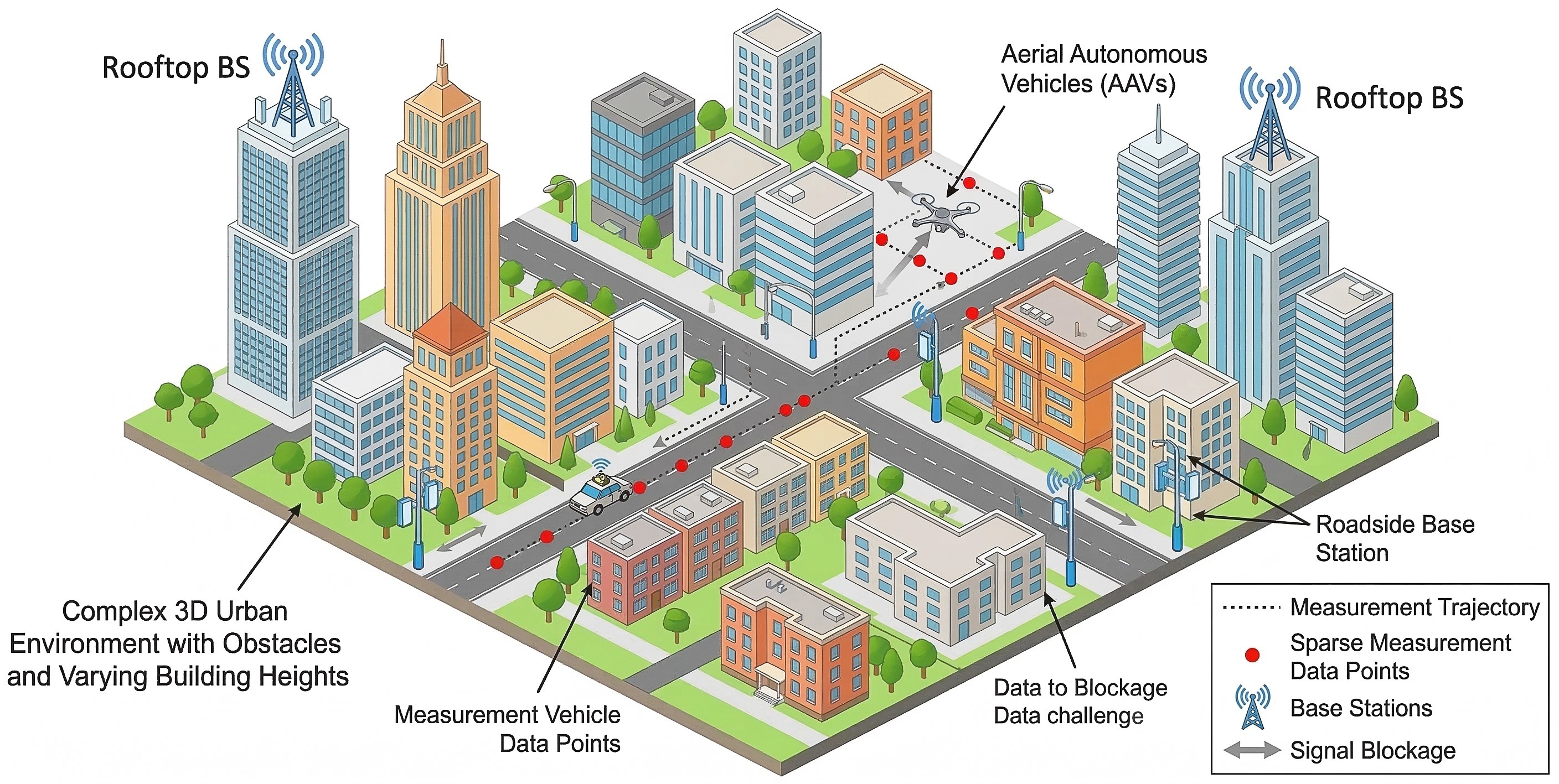}
    \caption{Radio Map reconstruction scenario from sparse measurements in a complex 3D urban environment.}
    \label{fig:architecture}
\end{figure*}

\section{System Model}
\label{sec_model}

\subsection{Network Scenario}
\label{subsec_network_scenario}

We consider a downlink wireless sensing and reconstruction scenario in a complex urban environment, as illustrated in Fig.~\ref{fig:architecture}. A transmitter, typically a base station (BS), is deployed at a known location $\mathbf{p}_{\mathrm{tx}} = [x_{\mathrm{tx}},y_{\mathrm{tx}},z_{\mathrm{tx}}]^{\mathsf T}$, and the objective is to infer a dense radio map over a target region of interest, following the propagation-map based formulation widely used in radio map and channel cartography studies \cite{Levie2021RadioUNet,Romero2024AerialBaseStationPlacementRadioMaps,Lee2017ChannelGainCartographyLowRankSparse}. Let $\mathcal{Q}\subset\mathbb{R}^{2}$ denote the horizontal service region. For tractable representation and learning, the region is discretized into an $H\times W$ grid. The center coordinate of the $(i,j)$-th grid cell and the corresponding dense gain radio map $\mathbf{G}$ are defined as:
\begin{align}
\mathbf{q}_{i,j} &= [x_{i,j},y_{i,j}]^{\mathsf T}\in\mathcal{Q}, \qquad 1\le i\le H,\;1\le j\le W, \label{eq:grid_coords}\\
\mathbf{G} &\in \mathbb{R}^{H\times W}, \label{eq:ckm_matrix}
\end{align}
where $G_{i,j}$ denotes the large-scale channel gain associated with location $\mathbf{q}_{i,j}$.

The environment contains buildings and other geometric structures that profoundly affect signal propagation. These deterministic scene attributes are encoded into several geometry maps, including obstacle occupancy, building height, relative transmitter position, transmitter distance, and a line-of-sight (LoS) proxy. Since measurements can only be collected in physically feasible outdoor regions, we further define an accessibility mask $\mathbf{M}_a\in\{0,1\}^{H\times W}$, where $M_{a,i,j}=1$ indicates that location $\mathbf{q}_{i,j}$ is accessible for measurement, while $M_{a,i,j}=0$ corresponds to inaccessible regions such as building interiors. This formulation is consistent with practical radio map acquisition, where the geographic support of the map, the physical scene structure, and the feasible sensing area are all known or can be precomputed from digital maps or ray-tracing based scene models \cite{Zhang2024RadiomapInpaintingRestrictedAreas,Chen2025Urban3DRadioMapSparse,Wang2025TowardsPreciseCKM}.

\subsection{Channel Gain Model}
\label{subsec_channel_gain_model}

The target quantity in this paper is the spatial distribution of channel gain. To support geometry-aware reconstruction, the physical gain model and the deterministic prior tensor $\mathbf{P}$ are formulated together as:
\begin{align}
G(\mathbf{q}) &= P_{\mathrm{tx}} - L(\mathbf{q}) - S(\mathbf{q}) + \varepsilon(\mathbf{q}), \label{eq:gain_model}\\
\mathbf{P} &= [\mathbf{O},\mathbf{H},\mathbf{R}_x,\mathbf{R}_y,\mathbf{D},\mathbf{L}], \label{eq:prior_tensor}
\end{align}
where for each location $\mathbf{q}\in\mathcal{Q}$, $P_{\mathrm{tx}}$ denotes the transmit power in dB scale, $L(\mathbf{q})$ represents the distance and geometry dependent path loss, $S(\mathbf{q})$ denotes shadowing caused by surrounding obstructions, and $\varepsilon(\mathbf{q})$ captures residual small-scale fluctuations or numerical approximation effects. In this work, the radio map is generated from ray-tracing simulation and mainly reflects the spatially structured propagation strength over the environment. Therefore, our reconstruction target should be interpreted primarily as a location dependent gain field dominated by large-scale propagation geometry, rather than instantaneous fast fading at a single carrier phase realization. After discretization, the continuous field $G(\mathbf{q})$ is represented by the grid map $\mathbf{G}$ defined above \cite{Lee2017ChannelGainCartographyLowRankSparse,Romero2024AerialBaseStationPlacementRadioMaps,Chen2025Urban3DRadioMapSparse}.

\begin{figure*}[t!]
\centering
\includegraphics[width=0.8\textwidth]{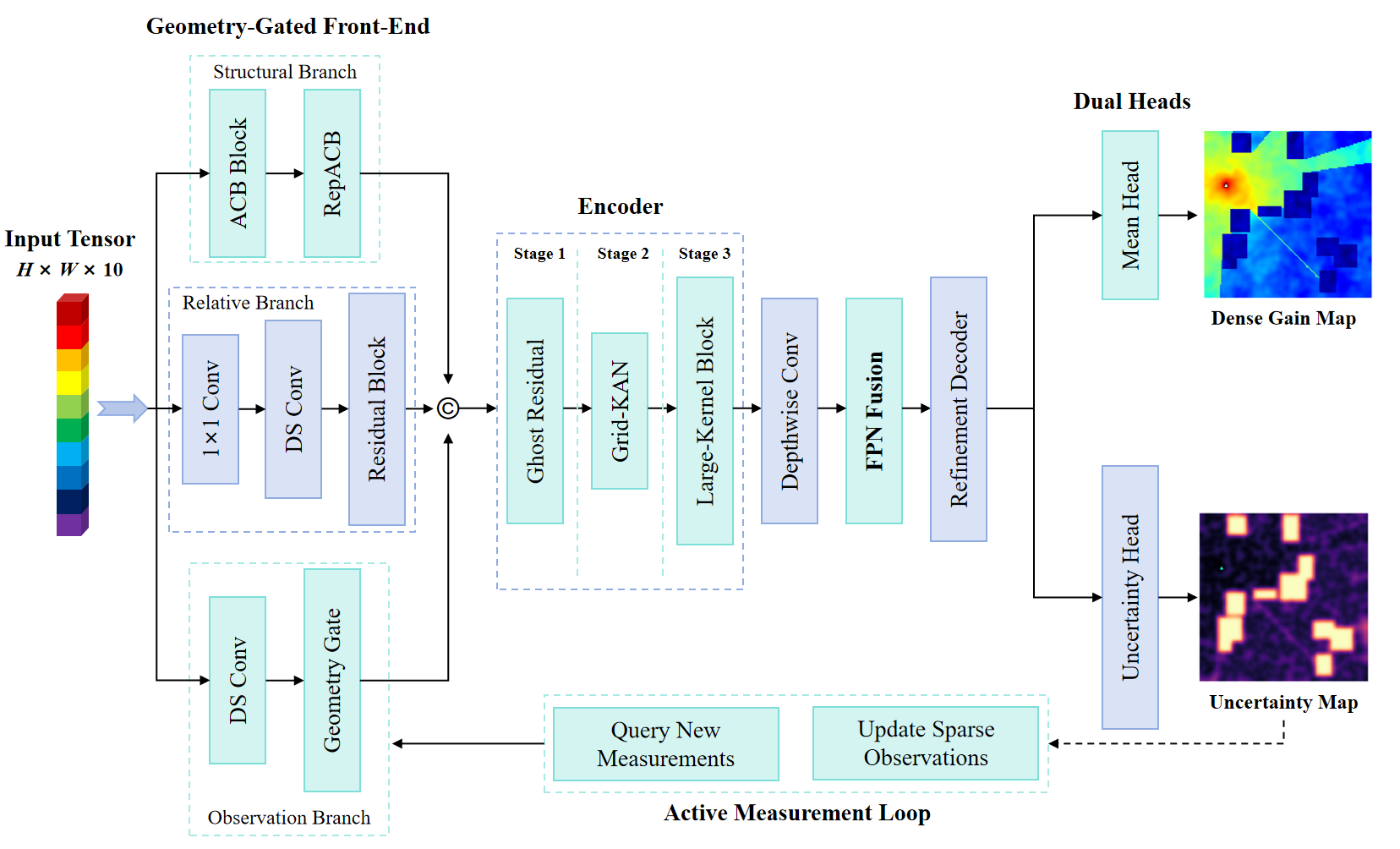}
\caption{The overall architecture of the proposed GeoUQ-GFNet. It integrates heterogeneous structural geometry cues, relative-position features, and sparse observations through a Geometry-Gated Front-End. Multi-scale representations are extracted using a Ghost and Grid-KAN-based encoder, followed by a feature pyramid network (FPN) fusion decoder. The dual heads produce both the reconstructed gain map and a predictive uncertainty map, which iteratively guides the active measurement loop.}
\label{fig:geouq_gfnet}
\end{figure*}

In the prior tensor representation \eqref{eq:prior_tensor}, $\mathbf{O}\in\{0,1\}^{H\times W}$ is the obstacle occupancy map, $\mathbf{H}\in\mathbb{R}^{H\times W}$ is the building height map, $\mathbf{R}_x$ and $\mathbf{R}_y\in\mathbb{R}^{H\times W}$ denote the relative horizontal coordinate maps with respect to the transmitter, $\mathbf{D}\in\mathbb{R}^{H\times W}$ is the transmitter-to-grid distance map, and $\mathbf{L}\in\{0,1\}^{H\times W}$ is a LoS proxy map indicating whether the straight line between the transmitter and a grid location is blocked by obstacles. These maps provide explicit geometric cues for learning propagation-aware spatial completion \cite{Levie2021RadioUNet,Zhang2024RadiomapInpaintingRestrictedAreas,Wang2025PointCloudCKM}.

\subsection{Measurement and Observation Model}
\label{subsec_measurement_model}

Because exhaustive measurement over all locations is costly, only a sparse subset of the accessible grid points can be observed. The observation model, the sparse gain map, and the resulting full network input are grouped as follows:
\begin{align}
\mathbf{Y} &= \mathbf{M}_s \odot (\mathbf{G} + \mathbf{N}), \label{eq:observation_model} \\
\mathbf{G}_s &= \mathbf{M}_s \odot \mathbf{G}, \label{eq:sparse_gain} \\
\mathbf{X} &= [\mathbf{P},\mathbf{G}_s,\mathbf{M}_s,\mathbf{M}_a,\mathbf{G}_{\mathrm{init}}], \label{eq:full_input}
\end{align}
where $\mathbf{M}_s\in\{0,1\}^{H\times W}$ denotes the sampling mask satisfying $\mathbf{M}_s \preceq \mathbf{M}_a$ ($M_{s,i,j}=1$ means a measurement is acquired at $\mathbf{q}_{i,j}$), $\odot$ denotes element-wise multiplication, and $\mathbf{N}\in\mathbb{R}^{H\times W}$ is the measurement noise map. In the simulation benchmark, the measurement noise is typically negligible compared with the propagation variation, so the observed sparse gain map is effectively computed as \eqref{eq:sparse_gain}. Furthermore, $\mathbf{G}_{\mathrm{init}}\in\mathbb{R}^{H\times W}$ represents a dense initialization map constructed from $(\mathbf{G}_s,\mathbf{M}_s,\mathbf{M}_a)$ by neighborhood-based iterative filling over unobserved accessible regions, which provides a coarse spatial prior to ease the prediction burden \cite{Teganya2022DeepCompletionAutoencoders,Jin2025I2IInpaintingCKM,Zhang2024RadiomapInpaintingRestrictedAreas}.

To study the impact of measurement support patterns, we consider three representative sparse sampling modes. The first is random sampling, where observed locations are randomly selected from accessible cells. The second is grid sampling, where measurements follow a structured lattice-like deployment. The third is road-constrained sampling, where measurements are limited to road-accessible trajectories. These modes produce different spatial support patterns even under the same sampling ratio and therefore induce different reconstruction difficulties \cite{Xu2024HowMuchDataCKM,Polyzos2024BayesianActiveLearningRadioMap,Shrestha2023SpectrumSurveying}.

\subsection{Reconstruction and Active Sensing Objectives}
\label{subsec_objectives}

Given the sparse observations and geometry-aware inputs defined in \eqref{eq:full_input}, the overall task studies sparse-to-dense radio map reconstruction and uncertainty-guided active measurement selection. The key mappings and spatial domain subsets are defined as:
\begin{align}
\hat{\mathbf{G}} &= f_{\theta}(\mathbf{X}), \label{eq:reconstruction_mapping}\\
\Omega_{\mathrm{unobs}} &= \{(i,j)\mid M_{a,i,j}=1,\;M_{s,i,j}=0\}, \label{eq:unobs_region}\\
\Omega_{\mathrm{cand}} &= \{(i,j)\mid M_{a,i,j}=1,\;M_{s,i,j}=0\}, \label{eq:cand_region}\\
\mathcal{Q}^{(t)} &\subseteq \Omega_{\mathrm{cand}}, \label{eq:query_set}
\end{align}
where $\hat{\mathbf{G}}$ is the reconstructed dense radio map, $\Omega_{\mathrm{unobs}}$ denotes the valid but currently unobserved region, $\Omega_{\mathrm{cand}}$ defines the candidate measurement region (which coincides with unobserved accessible locations), and $\mathcal{Q}^{(t)}$ denotes the subset of queried locations selected at iteration $t$.

The first objective is to minimize the prediction error over unobserved accessible locations \eqref{eq:unobs_region}, such that $\hat{G}_{i,j}\approx G_{i,j}$ for all $(i,j)\in\Omega_{\mathrm{unobs}}$. In addition to the mean gain map, the model also estimates a spatial predictive uncertainty map $\hat{\mathbf{U}}\in\mathbb{R}_{+}^{H\times W}$, reflecting the reconstruction confidence. Under a limited sensing budget, the second objective is to design an active sensing policy that logically selects the new measurement locations $\mathcal{Q}^{(t)}$ from $\Omega_{\mathrm{cand}}$ based on $\hat{\mathbf{U}}$, maximizing the overall reconstruction improvement under a fixed query budget \cite{Lee2019AdaptiveBayesianRadioTomography,Polyzos2024BayesianActiveLearningRadioMap,Shrestha2023SpectrumSurveying}.

\section{Methodology}
\label{sec_method}

\subsection{Overview of GeoUQ-GFNet}
\label{subsec_overview_method}

The overall architecture of the proposed GeoUQ-GFNet is illustrated in Fig.~\ref{fig:geouq_gfnet}. Given the input tensor $\mathbf{X}$ defined in \eqref{eq:full_input}, the network is designed to jointly predict a dense gain map and a spatial uncertainty map. The model consists of four main functional components. First, a Geometry-Gated Front-End separately encodes structural priors, relative position features, and sparse observations before executing adaptive fusion. Second, a lightweight multi-stage encoder based on Ghost modules, Grid-KAN blocks, and large-kernel convolutions manages multi-scale feature extraction. Third, a feature pyramid network (FPN)-based decoder intricately combines shallow details with deep semantic context. Finally, dual prediction heads jointly yield the gain reconstruction and the corresponding uncertainty estimation. Rather than directly regressing the final radio map, the model predicts a residual correction over the initialized map $\mathbf{G}_{\mathrm{init}}$, which stabilizes learning and encourages the network to focus on correcting initialization errors in difficult regions, consistent with deep completion and inpainting style reconstruction strategies \cite{Teganya2022DeepCompletionAutoencoders,Jin2025I2IInpaintingCKM,Zhang2024RadiomapInpaintingRestrictedAreas}.

\subsection{Geometry-Gated Front-End}
\label{subsec_geometry_gate}

To effectively integrate heterogeneous inputs, we first divide the input tensor into three semantically distinct groups. The structural geometry cues are denoted by $\mathbf{X}_{\mathrm{str}}=[\mathbf{O},\mathbf{H},\mathbf{M}_a,\mathbf{L},\mathbf{E}]$, where $\mathbf{E}$ represents an edge cue derived from obstacle and height discontinuities. The relative position features are organized as $\mathbf{X}_{\mathrm{rel}}=[\mathbf{R}_x,\mathbf{R}_y,\mathbf{D}]$, and the observation features are formed by $\mathbf{X}_{\mathrm{obs}}=[\mathbf{G}_s,\mathbf{M}_s,\mathbf{G}_{\mathrm{init}}]$. These three groups are encoded by dedicated branches as follows:
\begin{align}
\mathbf{F}_{\mathrm{str}} &= f_{\mathrm{str}}(\mathbf{X}_{\mathrm{str}}), \\
\mathbf{F}_{\mathrm{rel}} &= f_{\mathrm{rel}}(\mathbf{X}_{\mathrm{rel}}), \\
\mathbf{F}_{\mathrm{obs}} &= f_{\mathrm{obs}}(\mathbf{X}_{\mathrm{obs}}).
\end{align}
Here, $f_{\mathrm{str}}(\cdot)$ focuses on extracting geometry-aware structural features, $f_{\mathrm{rel}}(\cdot)$ captures large-scale spatial trends relative to the transmitter, and $f_{\mathrm{obs}}(\cdot)$ embeds the sparse measurement cues, in line with the broader environment-aware radio map modeling paradigm \cite{Levie2021RadioUNet,Li2024RadioGAT,Wang2025PointCloudCKM}.

\begin{figure}[t!]
\centering
\includegraphics[width=\columnwidth]{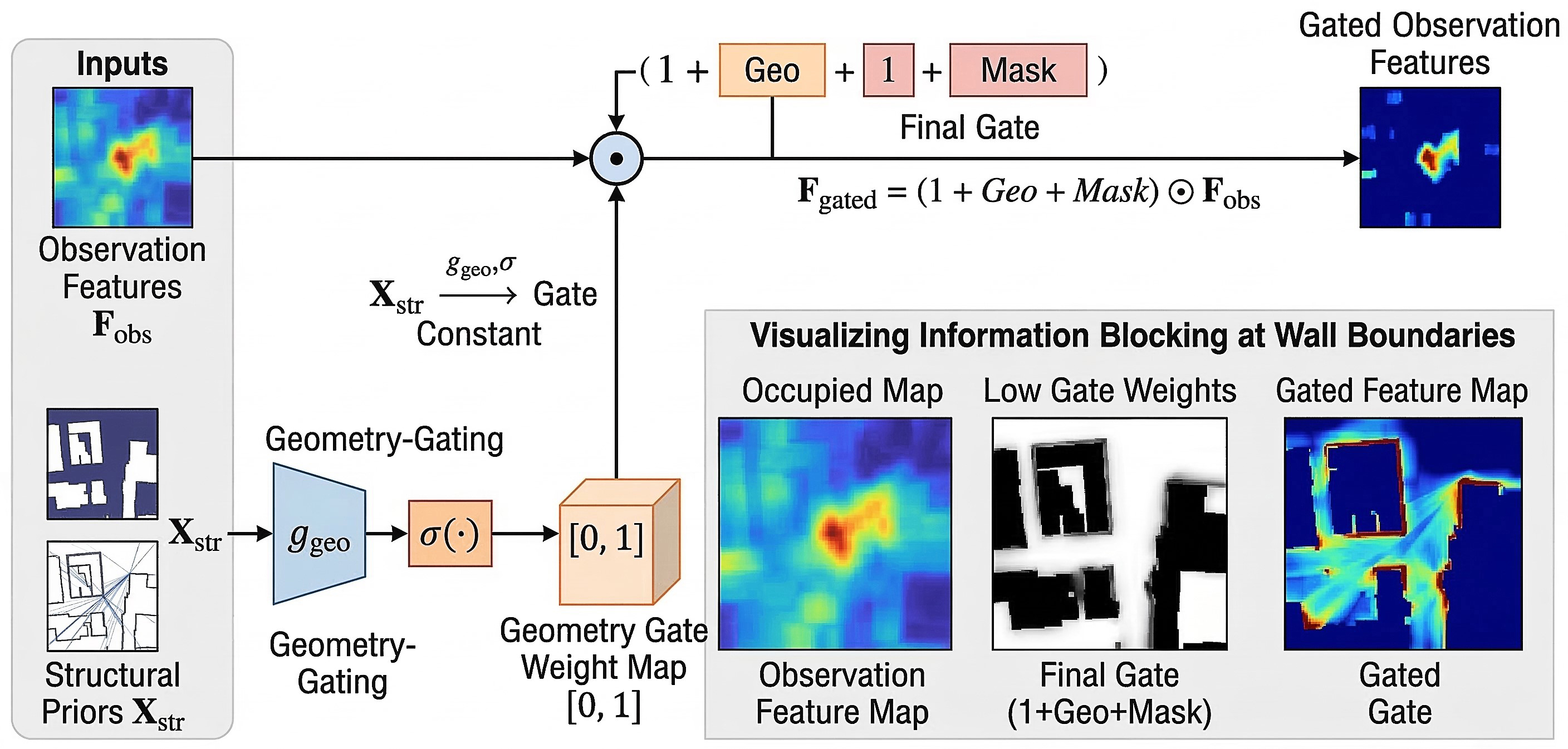}
\caption{Illustration of the Geometry-Gated Front-End. Observation features are dynamically modulated by geometry and mask gates derived from structural priors, explicitly blocking invalid information propagation at physical wall boundaries and inaccessible areas.}
\label{fig:geometry_gate}
\end{figure}

A key architectural design is that the observation branch is not fused directly but is dynamically modulated by geometry dependent and mask dependent gates. These gates and the resulting modulated observation feature are computed as
\begin{align}
\mathbf{G}_{\mathrm{geo}} &= g_{\mathrm{geo}}(\mathbf{X}_{\mathrm{str}}), \\
\mathbf{G}_{\mathrm{mask}} &= g_{\mathrm{mask}}([\mathbf{M}_s,\mathbf{M}_a]), \\
\tilde{\mathbf{F}}_{\mathrm{obs}} &= \mathbf{F}_{\mathrm{obs}} \odot \left( 1+\mathbf{G}_{\mathrm{geo}}+\mathbf{G}_{\mathrm{mask}} \right).
\end{align}
This mechanism explicitly suppresses invalid or misleading propagation across obstacles and inaccessible regions while preserving informative cues from observed locations, which is particularly relevant to restricted-area radiomap reconstruction and geometry-guided completion \cite{Zhang2024RadiomapInpaintingRestrictedAreas,Chen2025Urban3DRadioMapSparse}. Finally, the three branches are fused as $\mathbf{F}_0 = f_{\mathrm{fuse}} \!\left( [\mathbf{F}_{\mathrm{str}},\mathbf{F}_{\mathrm{rel}},\tilde{\mathbf{F}}_{\mathrm{obs}}] \right)$, where $f_{\mathrm{fuse}}(\cdot)$ is implemented using $1\times1$ projection and residual refinement.

\subsection{Lightweight Multi-Stage Encoder}
\label{subsec_encoder}

Built upon the fused feature $\mathbf{F}_0$, the backbone follows a progressive encoder context fusion design. The encoder contains three cascaded stages configured as
\begin{align}
(\mathbf{F}_1,\mathbf{S}_1) &= \mathcal{E}_1(\mathbf{F}_0), \\
(\mathbf{F}_2,\mathbf{S}_2) &= \mathcal{E}_2(\mathbf{F}_1), \\
(\mathbf{F}_3,\mathbf{S}_3) &= \mathcal{E}_3(\mathbf{F}_2),
\end{align}
where $\mathbf{S}_1$, $\mathbf{S}_2$, and $\mathbf{S}_3$ are multi-scale features safely retained for the subsequent decoder fusion. To improve efficiency, the encoder mainly adopts Ghost residual blocks. For an input feature map $\mathbf{Z}$, the Ghost residual transformation is abstractly expressed as $\mathcal{G}(\mathbf{Z}) = \phi\!\left( \mathbf{Z}+h_{\mathrm{ghost}}(\mathbf{Z}) \right)$, where $h_{\mathrm{ghost}}(\cdot)$ denotes lightweight feature generation and $\phi(\cdot)$ is the nonlinear activation. Compared with standard convolution blocks, this design significantly reduces feature redundancy while thoroughly maintaining adequate representation capability \cite{Han2020GhostNet}. The overall encoder--decoder organization also follows the practical design philosophy of modern deep radio map reconstruction networks \cite{Levie2021RadioUNet,Teganya2022DeepCompletionAutoencoders,Liu2026DATUnetRadioMap}.

\subsection{Grid-KAN Nonlinear Enhancement}
\label{subsec_gridkan}

To better model complex nonlinearity induced by blockage transitions, shadow regions, and distance dependent attenuation, Grid-KAN blocks are logically inserted into the deeper encoder stages. 
\begin{figure}[t!]
\centering
\includegraphics[width=\columnwidth]{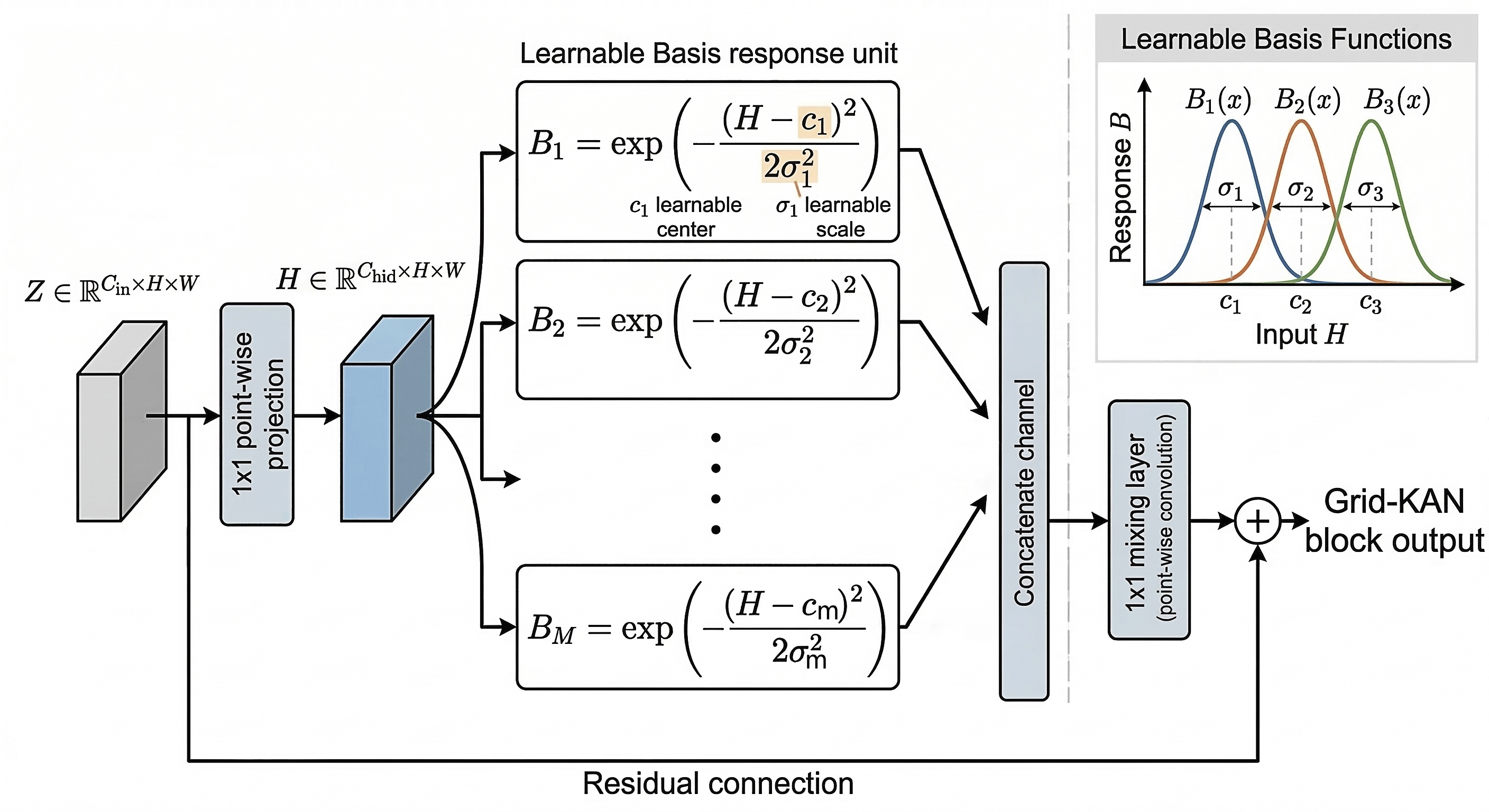}
\caption{Detailed architecture of the Grid-KAN block. Intermediate features are projected to a hidden dimension and processed by multiple learnable radial basis functions, whose outputs are concatenated and mixed to enhance the nonlinear fitting capability.}
\label{fig:grid_kan}
\end{figure}
Given an intermediate feature map $\mathbf{Z}$, the Grid-KAN layer first projects it to a hidden representation $\mathbf{H}$ and then applies a set of learnable radial basis responses parameterized as
\begin{equation}
\mathbf{B}_m = \exp\!\left( -\frac{(\mathbf{H}-c_m)^2}{2\sigma_m^2} \right), \qquad m=1,\ldots,M,
\end{equation}
where $c_m$ and $\sigma_m$ are learnable centers and scales, and $M$ is the strictly defined number of basis functions. These responses are organically concatenated and mixed to form the output feature $\mathcal{K}(\mathbf{Z}) = \psi\!\left( [\mathbf{B}_1,\ldots,\mathbf{B}_M] \right)$, where $\psi(\cdot)$ denotes comprehensive channel mixing and projection. In our overarching design, Grid-KAN is combined with residual and depthwise separable convolutional operations, enhancing the ability of the backbone to fit irregular channel variation patterns that are inherently difficult to capture by plain convolution alone \cite{Liu2025KAN,Liao2026RadioKAN}.

\subsection{Large-Kernel Context Modeling}
\label{subsec_large_kernel}

Radio map reconstruction requires both localized boundary awareness and a much broader spatial context. Therefore, the deepest encoder stage and bottleneck context module employ large-kernel depthwise convolution to robustly enlarge the receptive field. The context feature is mathematically written as $\mathbf{F}_{\mathrm{ctx}}=\mathcal{C}(\mathbf{F}_3)$, where $\mathcal{C}(\cdot)$ combines large-kernel depthwise convolution, small-kernel refinement, channel mixing, and residual connection. This specific design is highly suitable for radio map completion because large-scale signal trends are primarily governed by broad geometry layout rather than purely local texture continuity, which is consistent with recent deep reconstruction and attention-based radiomap estimation studies \cite{Levie2021RadioUNet,Liu2026DATUnetRadioMap,Chen2025Urban3DRadioMapSparse}.

\subsection{FPN-Based Multi-Scale Fusion}
\label{subsec_fpn}

To merge shallow structural details and deep semantic features efficiently, we adopt a feature pyramid network (FPN) style fusion module. The multi-scale features are first projected to a common channel dimension, aligned spatially, and further refined through the following operations:
\begin{align}
\mathbf{F}_{\mathrm{fpn}} &= f_{\mathrm{fpn}} \!\left( \left[ p_1(\mathbf{S}_1), \mathrm{Up}\!\left(p_2(\mathbf{S}_2)\right), \mathrm{Up}\!\left(p_3(\mathbf{F}_{\mathrm{ctx}})\right) \right] \right), \\
\mathbf{F}_{\mathrm{ref}} &= f_{\mathrm{ref}}(\mathbf{F}_{\mathrm{fpn}}).
\end{align}
Here, $p_1(\cdot)$, $p_2(\cdot)$, and $p_3(\cdot)$ are lateral projections, $\mathrm{Up}(\cdot)$ denotes standard bilinear upsampling, and $\mathbf{F}_{\mathrm{ref}}$ serves as the shared representation for the final prediction heads. This multi-scale fusion strategy is also compatible with prior radio map estimation architectures that rely on hierarchical spatial features \cite{Levie2021RadioUNet,Teganya2022DeepCompletionAutoencoders}.

\subsection{Residual Gain Reconstruction}
\label{subsec_residual_reconstruction}

Instead of directly predicting the full gain map, GeoUQ-GFNet estimates a targeted residual correction over the initialization map $\mathbf{G}_{\mathrm{init}}$, and conditionally merges it to form the final reconstruction map. The progression of these mapping transformations is expressed as
\begin{align}
\Delta\mathbf{G} &= f_{\mathrm{mean}}(\mathbf{F}_{\mathrm{ref}}), \\
\hat{\mathbf{G}}_{u} &= \mathbf{G}_{\mathrm{init}}+\Delta\mathbf{G}, \\
\hat{\mathbf{G}} &= \mathbf{M}_s\odot\mathbf{G}_s + (1-\mathbf{M}_s)\odot\hat{\mathbf{G}}_{u}. \label{eq:final_reconstruction_method}
\end{align}
In practice, the prediction can be optionally restricted to a physically reasonable gain range during implementation, but the core analytical formulation is strictly given by \eqref{eq:final_reconstruction_method}. This residual strategy is highly advantageous because the initialization already captures coarse spatial continuity, allowing the network to concentrate strictly on correcting hard regions such as obstacle boundaries, non-line-of-sight (NLoS) transitions, and sparsely supported areas \cite{Teganya2022DeepCompletionAutoencoders,Jin2025I2IInpaintingCKM,Zhang2024RadiomapInpaintingRestrictedAreas}.

\subsection{Uncertainty Estimation}
\label{subsec_uncertainty}

To thoroughly quantify predictive confidence, GeoUQ-GFNet uses a second prediction head to estimate a spatial log-variance map. Specifically, the uncertainty branch conditions the shared feature on several observation-related masks, yielding the final predictive standard deviation map through the operations:
\begin{align}
\mathbf{F}_{\mathrm{uq}} &= f_{\mathrm{uq}} \!\left( [\mathbf{F}_{\mathrm{ref}},\mathbf{M}_s,\mathbf{M}_a,\mathbf{L}] \right), \\
\mathbf{S} &= f_{\mathrm{var}} \!\left( \mathcal{K}(\mathbf{F}_{\mathrm{uq}}) \right), \quad \text{where} \;\; \mathbf{S}=\log \hat{\boldsymbol{\sigma}}^2, \\
\hat{\mathbf{U}} &= \exp\!\left(\frac{1}{2}\mathbf{S}\right). \label{eq:uncertainty_map_method}
\end{align}
Hence, for each location $(i,j)$, the model safely outputs $\hat{G}_{i,j}=\mu_{i,j}$ and $s_{i,j}=\log \hat{\sigma}_{i,j}^{2}$. We actively adopt a heteroscedastic Gaussian likelihood over the unobserved accessible region, modeled as
\begin{equation}
\mathcal{L}_{\mathrm{NLL}} = \sum_{(i,j)\in\Omega_{\mathrm{unobs}}} \left[ \frac{1}{2}e^{-s_{i,j}}(\hat{G}_{i,j}-G_{i,j})^{2} + \frac{1}{2}s_{i,j} \right].
\end{equation}
This formulation allows the model to reliably assign larger uncertainty to difficult regions that exhibit intrinsically higher reconstruction ambiguity, which is aligned with the uncertainty-aware radio map sampling literature \cite{Lee2019AdaptiveBayesianRadioTomography,Polyzos2024BayesianActiveLearningRadioMap}.

\subsection{Training Objective}
\label{subsec_training_objective}

To balance reconstruction fidelity and uncertainty quality seamlessly, the full training objective incorporates four core terms into the total loss calculation $\mathcal{L} = \lambda_{1}\mathcal{L}_{1} + \lambda_{\mathrm{grad}}\mathcal{L}_{\mathrm{grad}} + \lambda_{\mathrm{nll}}\mathcal{L}_{\mathrm{NLL}} + \lambda_{\mathrm{var}}\mathcal{L}_{\mathrm{reg}}$. Here, $\mathcal{L}_{1}$ is the masked $\ell_1$ reconstruction loss isolated precisely over $\Omega_{\mathrm{unobs}}$, $\mathcal{L}_{\mathrm{grad}}$ is a gradient consistency loss that encourages continuous local spatial smoothness alongside edge preservation, $\mathcal{L}_{\mathrm{NLL}}$ is the robust heteroscedastic Gaussian negative log-likelihood detailed previously, and $\mathcal{L}_{\mathrm{reg}}$ is an essential variance regularization term strategically used to stabilize the uncertainty estimation. This unified objective effectively encourages the network to produce both accurate dense radio map estimates and uncertainty maps that remain highly meaningful for downstream active sensing scenarios \cite{Teganya2022DeepCompletionAutoencoders,Polyzos2024BayesianActiveLearningRadioMap}.

\begin{algorithm}[htbp]
\caption{Radio Map Completion and Active Sensing}
\label{alg:overall_method}
\begin{algorithmic}[1]
\REQUIRE Geometry maps $\{\mathbf{O},\mathbf{H},\mathbf{R}_x,\mathbf{R}_y,\mathbf{D},\mathbf{L},\mathbf{M}_a\}$, initial sparse observations $(\mathbf{G}_s^{(0)},\mathbf{M}^{(0)})$, query budget $K$, active iteration number $T$, trained network $\mathcal{F}_{\theta}$
\ENSURE Final reconstructed gain map $\hat{\mathbf{G}}^{(T)}$

\STATE $\mathbf{G}_{\mathrm{init}}^{(0)} \leftarrow \mathrm{Init}(\mathbf{G}_s^{(0)},\mathbf{M}^{(0)},\mathbf{M}_a)$

\FOR{$t=0$ \TO $T-1$}
    \STATE $\mathbf{X}^{(t)} \leftarrow [\mathbf{O},\mathbf{H},\mathbf{R}_x,\mathbf{R}_y,\mathbf{D},\mathbf{L},\mathbf{G}_s^{(t)},\mathbf{M}^{(t)},\mathbf{M}_a,\mathbf{G}_{\mathrm{init}}^{(t)}]$
    \STATE $(\Delta\hat{\mathbf{G}}^{(t)}, \mathbf{S}^{(t)}) \leftarrow \mathcal{F}_{\theta}(\mathbf{X}^{(t)})$
    \STATE $\hat{\mathbf{G}}_{u}^{(t)} \leftarrow \mathbf{G}_{\mathrm{init}}^{(t)}+\Delta\hat{\mathbf{G}}^{(t)}$
    \STATE $\hat{\mathbf{G}}^{(t)} \leftarrow \mathbf{M}^{(t)}\odot\mathbf{G}_s^{(t)} + (1-\mathbf{M}^{(t)})\odot\hat{\mathbf{G}}_{u}^{(t)}$
    \STATE $\hat{\mathbf{U}}^{(t)} \leftarrow \exp\!\left(\frac{1}{2}\mathbf{S}^{(t)}\right)$
    \STATE $\Omega^{(t)} \leftarrow \{(i,j)\mid M_{a,i,j}=1,\;M_{i,j}^{(t)}=0\}$
    \IF{$\Omega^{(t)}=\emptyset$}
        \STATE \textbf{break}
    \ENDIF
    \STATE $\mathcal{Q}^{(t)} \leftarrow \operatorname{TopK}\Big(\hat{\mathbf{U}}^{(t)}\odot\mathbf{M}_a\odot(1-\mathbf{M}^{(t)})\Big)$
    \STATE $\mathbf{M}^{(t+1)} \leftarrow \mathbf{M}^{(t)} \cup \mathcal{Q}^{(t)}$
    \STATE $\mathbf{G}_s^{(t+1)} \leftarrow \mathbf{M}^{(t+1)}\odot\mathbf{G}$
    \STATE $\mathbf{G}_{\mathrm{init}}^{(t+1)} \leftarrow \mathrm{Init}(\mathbf{G}_s^{(t+1)},\mathbf{M}^{(t+1)},\mathbf{M}_a)$
\ENDFOR
\STATE \RETURN $\hat{\mathbf{G}}^{(T)}$
\end{algorithmic}
\end{algorithm}

\subsection{Uncertainty-Guided Active Measurement Selection}
\label{subsec_active_sampling}

Beyond simple one-shot reconstruction, we fundamentally consider an iterative sensing setting in which the model can actively request additional measurements under a tightly limited budget. At iteration $t$, the current observation mask is denoted by $\mathbf{M}^{(t)}$, and the model predicts both a reconstructed radio map $\hat{\mathbf{G}}^{(t)}$ and a targeted uncertainty map $\hat{\mathbf{U}}^{(t)}$. Candidate query locations are rigorously restricted to accessible but unobserved cells forming the set $\Omega^{(t)} = \{(i,j)\mid M^{(t)}_{i,j}=0,\;M_{a,i,j}=1\}$. The next optimal query set is smartly selected by ranking the uncertainty values over this candidate region using $\mathcal{Q}^{(t)} = \operatorname{TopK} \Big( \hat{\mathbf{U}}^{(t)}\odot\mathbf{M}_a\odot(1-\mathbf{M}^{(t)}) \Big)$, where $\operatorname{TopK}(\cdot)$ returns the $K$ highest-scoring candidate locations. After successfully querying these target locations, the sparse measurement map and observation mask are functionally updated as
\begin{align}
\mathbf{M}^{(t+1)} &= \mathbf{M}^{(t)} \cup \mathcal{Q}^{(t)}, \\
\mathbf{G}_s^{(t+1)} &= \mathbf{M}^{(t+1)}\odot \mathbf{G}.
\end{align}
Following this collection, the fundamental initialization map $\mathbf{G}_{\mathrm{init}}^{(t+1)}$ is rapidly recomputed from the newly updated observations, and the model is seamlessly re-applied for the following iteration. The complete iterative procedure is summarized in Algorithm~\ref{alg:overall_method}. Since remarkably high uncertainty usually appears in poorly supported or geometrically difficult regions, this intuitive policy naturally prioritizes acquiring measurements that are mathematically expected to improve the final radio map significantly more efficiently than uninformed random selection methodologies \cite{Lee2019AdaptiveBayesianRadioTomography,Polyzos2024BayesianActiveLearningRadioMap,Shrestha2023SpectrumSurveying}.


\section{Experimental Setup}
\label{sec_experiments}

\subsection{Dataset Generation}
\label{subsec_dataset_generation}

The experiments are conducted on the proposed UrbanRT-RM, a controlled ray tracing benchmark generated from synthetic urban scenes. In total, we construct seven scene instances, including four road oriented layouts and three non road building layouts. The road oriented group contains a crossroad scene, a T junction scene, a canyon scene, and an offset crossroad scene. The non road group contains dense, medium, and sparse building layouts. This design allows us to jointly vary road topology, building density, and spatial blockage complexity, while also complementing recent benchmark efforts on AI-based CKM and radio map learning \cite{Wu2025CKMImageNet,Zhang2025SpectrumNet}.

\begin{figure}[t]
    \centering
    \begin{subfigure}[t]{0.48\linewidth}
        \centering
        \includegraphics[height=2.6cm, keepaspectratio]{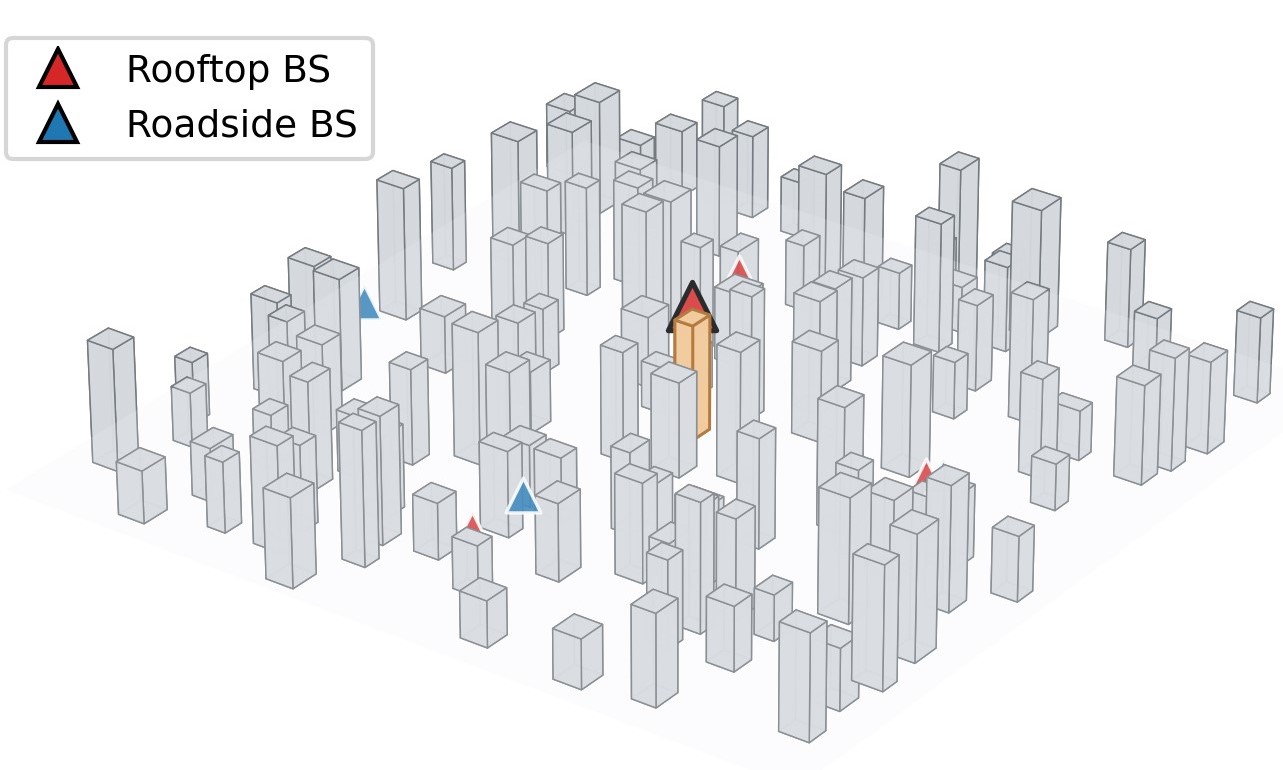}
        \caption{Mixed rooftop and roadside BSs deployment modes.}
        \label{fig:dataset_bs_types}
    \end{subfigure}
    \hfill
    \begin{subfigure}[t]{0.48\linewidth}
        \centering
        \includegraphics[height=3.4cm, keepaspectratio]{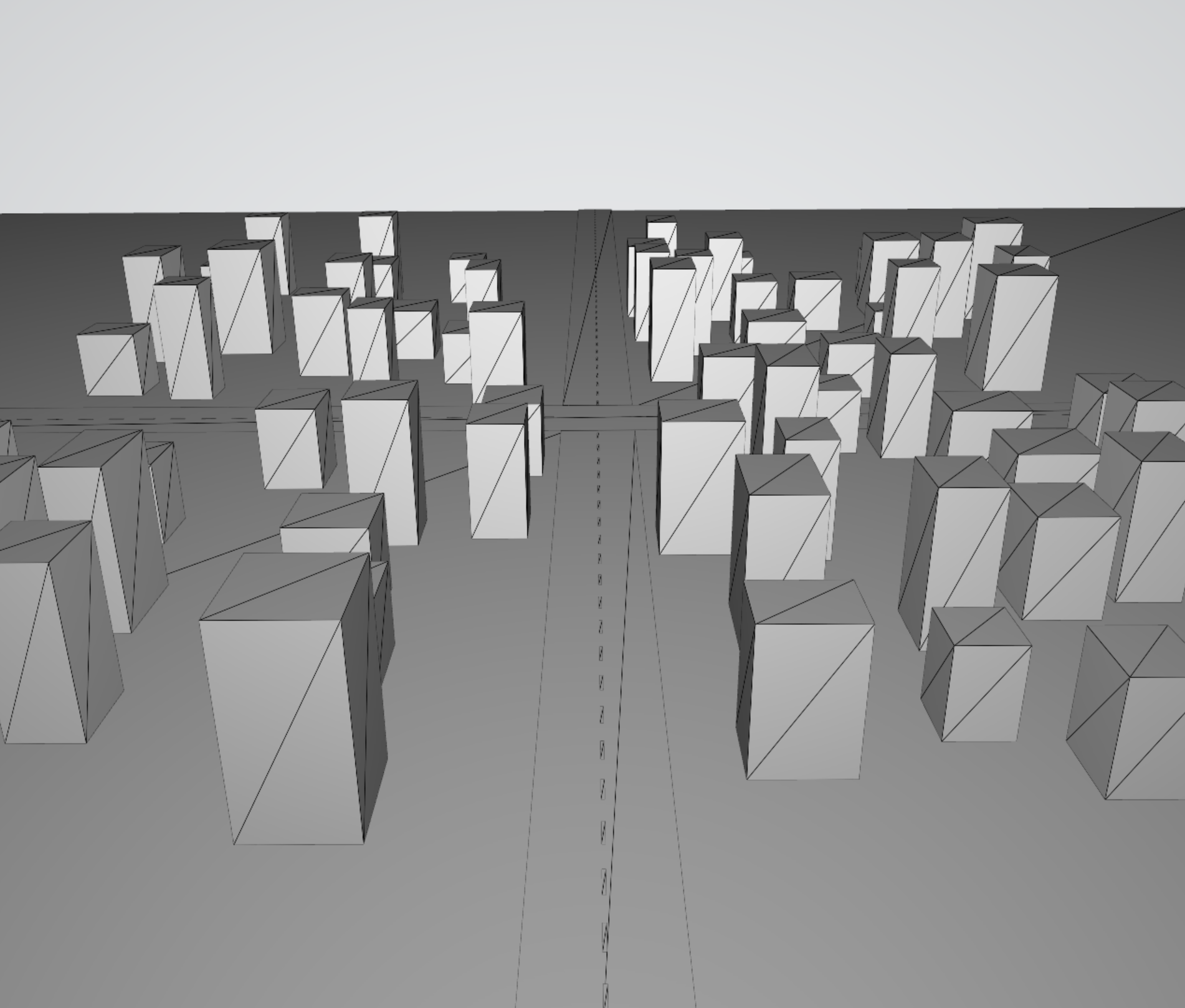}
        \caption{3D perspective view of a synthetic urban scene used for RT.}
        \label{fig:dataset_scene_3d}
    \end{subfigure}

    \vspace{0.6em}

    \begin{subfigure}[t]{0.48\linewidth}
        \centering
        \includegraphics[height=3.6cm, keepaspectratio]{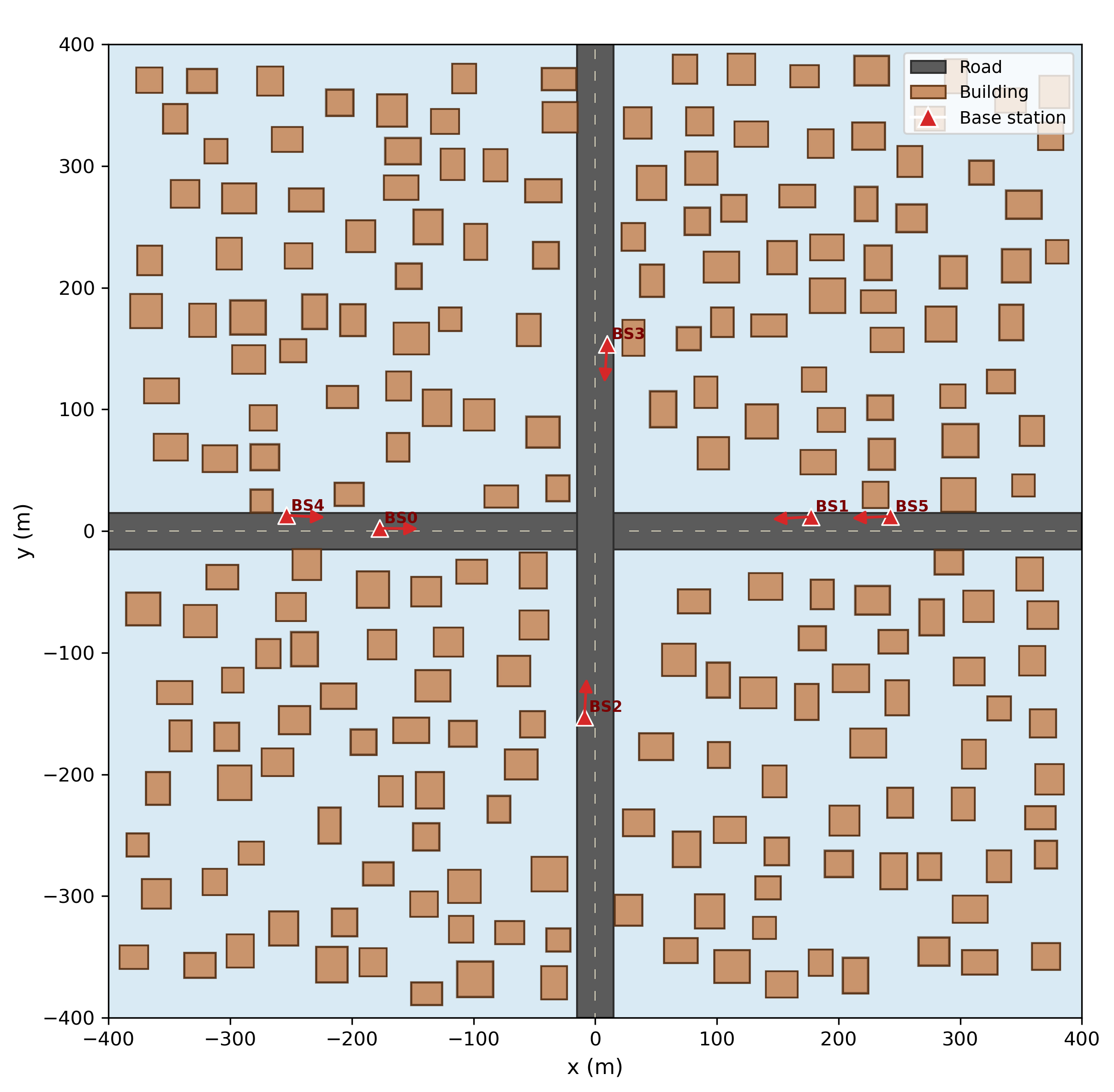}
        \caption{2D scene layout with road geometry, buildings, and BS.}
        \label{fig:dataset_scene_2d}
    \end{subfigure}
    \hfill
    \begin{subfigure}[t]{0.48\linewidth}
        \centering
        \includegraphics[height=3.6cm, keepaspectratio]{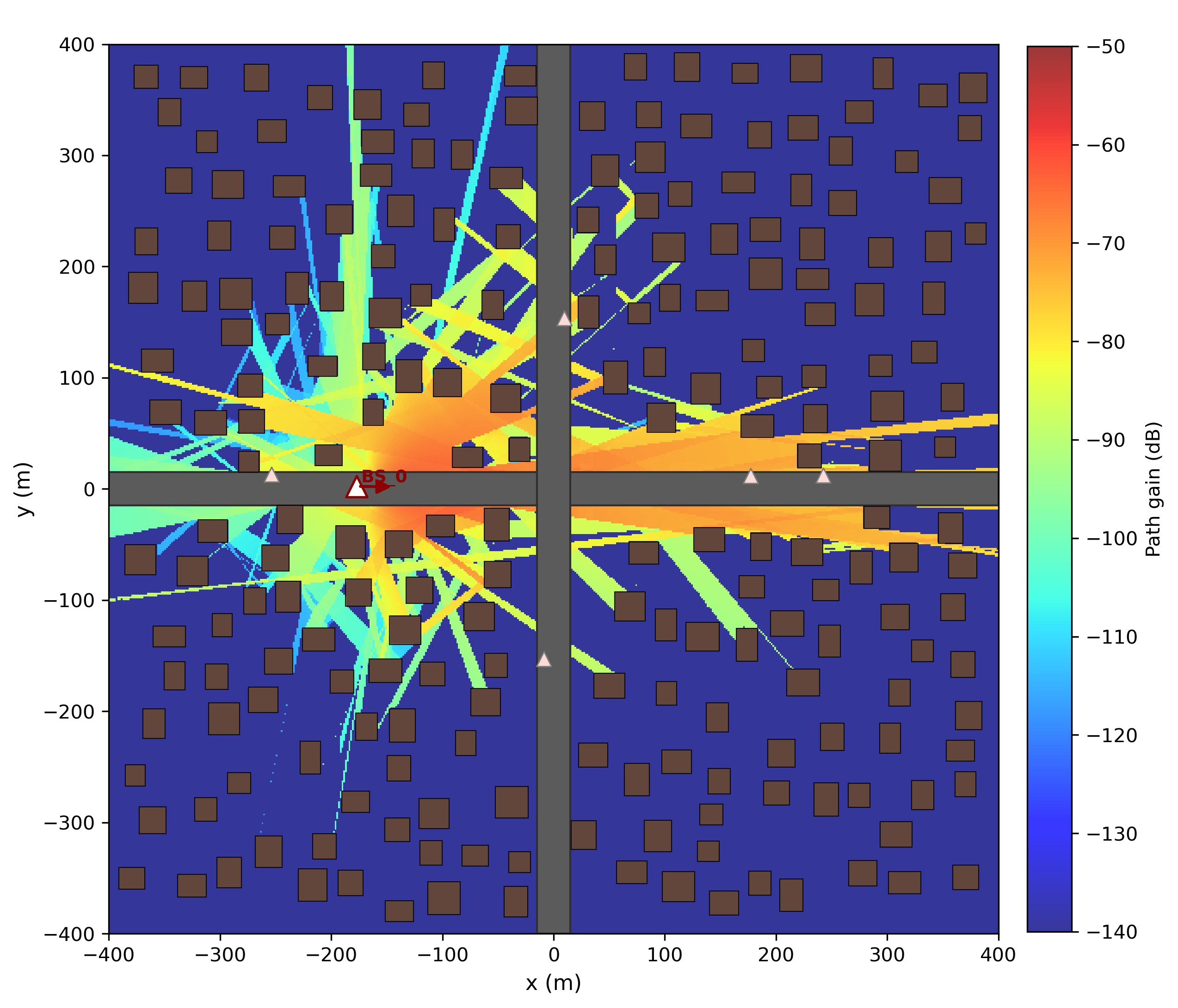}
        \caption{Dense gain map generated by the RT simulator.}
        \label{fig:dataset_gainmap}
    \end{subfigure}
    \caption{Illustration of the proposed UrbanRT-RM synthetic urban ray tracing dataset. The dataset includes diverse 2D road/building layouts, corresponding 3D geometric scene realizations, mixed rooftop/roadside BS deployments, and dense gain maps generated by Sionna RT. Together, these components provide paired geometry and propagation data for radio map reconstruction under different blockage and sampling conditions.}
    \label{fig:dataset_overview}
\end{figure}

For each scene, buildings are randomly generated under geometric constraints on width, length, height, and inter building clearance. More specifically, the map extent is set to $400 \text{ m}$ on each horizontal axis, the global map resolution is $512 \times 512$, and the cropped patch size is $128 \times 128$. The street width is fixed to $30 \text{ m}$. Depending on the scene category, the number of buildings varies from $120$ to $300$, producing low density, medium density, and high density propagation environments. Each scene contains $8$ BS deployments drawn from roadside and rooftop modes, which yields diverse transmitter positions and viewing geometries. Representative examples of the generated layouts, 3D scene realizations, deployment modes, and ray traced gain fields are shown in Fig.~\ref{fig:dataset_overview}. Such a controlled scene-and-deployment design is also motivated by recent interest in standardized CKM/radio-map datasets and urban sparse-measurement evaluation \cite{Wu2025CKMImageNet,Zhang2025SpectrumNet,Chen2025Urban3DRadioMapSparse}.

Dense gain maps are generated by the Sionna RT engine under a unified propagation configuration \cite{hoydis2023sionnart}. The carrier frequency is set to $3.5$ GHz, the receiver height is set to $1.5$ m, and the ray tracing depth is limited to $3$. For each transmitter, the ray tracing solver uses $2 \times 10^{5}$ samples to compute the dense gain field over the full spatial grid. The resulting target map therefore represents a location dependent large scale gain field shaped by distance attenuation, shadowing, and geometric blockage, which is consistent with the propagation-map abstraction adopted in recent radio map and CKM studies \cite{Levie2021RadioUNet,Romero2024AerialBaseStationPlacementRadioMaps,Chen2025Urban3DRadioMapSparse}.

Based on the generated scene and BS configuration, we further construct geometry aware side information channels, including the obstacle map, height map, relative horizontal coordinates, transmitter distance, accessible region mask, and LoS proxy map. The final reconstruction input also includes sparse observations and a dense initialization obtained from iterative neighborhood propagation, following the general idea of combining environment priors and sparse observations in deep radio map reconstruction \cite{Levie2021RadioUNet,Zhang2024RadiomapInpaintingRestrictedAreas,Wang2025TowardsPreciseCKM}.

To evaluate the effect of measurement support, we generate sparse observations under three sampling modes, namely random sampling, grid sampling, and road constrained sampling. The observation ratio is selected from $\{5\%, 10\%, 20\%, 40\%\}$. In the main comparison, we focus on the random $10\%$ setting, while additional experiments are conducted under random $5\%$, grid $10\%$, and road $10\%$ conditions. For each valid BS, we extract $160$ patches, subject to constraints on building coverage and valid gain support, so that each patch contains sufficient structural diversity for reconstruction. The choice to explicitly vary the observation ratio and support pattern is also related to prior discussions on sample efficiency and data requirement in CKM/radio map construction \cite{Xu2024HowMuchDataCKM,Polyzos2024BayesianActiveLearningRadioMap}.

The final dataset index is built at the patch level, and the train, validation, and test splits are generated using non overlapping patch identities. The split ratios are $70\%$, $15\%$, and $15\%$, respectively. Since the split is performed by patch identity rather than by sampling variant, different sparse versions of the same underlying patch never appear across different data partitions. This prevents information leakage and ensures a fair evaluation protocol, which is consistent with benchmark-oriented evaluation practice in recent radio map and CKM datasets \cite{Wu2025CKMImageNet,Zhang2025SpectrumNet}.

\subsection{Implementation Details}
\label{subsec_implementation_details}

\subsubsection{Hardware and Software Environment}

All experiments were conducted on a workstation equipped with an Intel Core i9-12900KF CPU, an NVIDIA GeForce RTX 5060 Ti GPU with 16 GB memory, and 64 GB RAM. The ray-tracing benchmark was generated in Python 3.11 using Sionna RT together with TensorFlow, while the reconstruction models were implemented in PyTorch 2.x. GPU acceleration was enabled through CUDA 11.8 and cuDNN. The dataset generation pipeline also uses the Sionna RT backend.

\subsubsection{Training Strategy}

The proposed GeoUQ-GFNet is trained for the radio map completion task with the random $10\%$ sparse observation setting as the default training variant. Unless otherwise stated, the model uses a batch size of $8$ and is optimized for at most $120$ epochs using AdamW with learning rate $10^{-4}$ and weight decay $10^{-5}$. A ReduceLROnPlateau scheduler is used to reduce the learning rate by a factor of $0.5$ when the validation loss saturates. Early stopping is applied with a patience of $20$ epochs. To stabilize optimization, gradient clipping with maximum norm $1.0$ is used throughout training.
The GeoUQ-GFNet configuration follows the implementation in our codebase. The base channel width is set to $32$, the hidden dimension of the Grid-KAN module is set to $32$, the Ghost ratio is set to $2$, the FPN channel dimension is set to $64$, and the number of radial basis functions in each Grid-KAN layer is set to $10$. The dropout rate is set to zero in the reported experiments. The lightweight backbone design is motivated by the efficiency-oriented feature generation of Ghost modules and the nonlinear modeling capability of KAN-style representations \cite{Han2020GhostNet,Liu2025KAN,Liao2026RadioKAN}.
Input normalization is performed channel-wise. The height map is normalized by the maximum height value observed in the training set. The relative coordinate maps and distance map are standardized using the training-set mean and standard deviation. Gain-related channels, including sparse gain and initialization gain, are clipped to the interval from $-140$ dB to $-50$ dB and then linearly normalized to $[0,1]$. The target gain map is normalized in the same way.
To improve generalization, data augmentation is applied only to the training set. Specifically, random rotations and horizontal or vertical flips are used. During augmentation, the relative coordinate channels are transformed consistently with the geometric operation so that their physical meaning is preserved after rotation or reflection.
For the composite training objective defined in Section~\ref{sec_method}, the empirical loss weights are set to $\lambda_{1}=1.0$, $\lambda_{\mathrm{grad}}=0.05$, $\lambda_{\mathrm{nll}}=0.2$, and $\lambda_{\mathrm{var}}=0.001$. The reconstruction and negative log-likelihood terms are computed exclusively over unobserved accessible cells, whereas the gradient consistency term is evaluated over all valid accessible regions. To maintain numerical stability during optimization, the predicted log-variance map from the uncertainty head is explicitly clipped to the range of $[-6.0, 2.0]$.
For active measurement evaluation, the trained model is iteratively queried on test patches. The initial sparse observation is updated over $4$ rounds, and each round adds an additional observation budget equal to $1\%$ of the patch area. The uncertainty-guided policy selects the highest-uncertainty unobserved accessible cells, while the random baseline uses the same budget but samples uniformly from the candidate set. This evaluation protocol is closely related to prior active radio map acquisition works that compare uncertainty-driven querying against uninformed sampling under matched budgets \cite{Lee2019AdaptiveBayesianRadioTomography,Polyzos2024BayesianActiveLearningRadioMap,Shrestha2023SpectrumSurveying}.

\begin{table*}[!t]
\centering
\caption{Scene level average RMSE in dB}
\label{tab:scene_level_rmse}
\small
\setlength{\tabcolsep}{10pt}
\begin{tabular}{lccccc}
\toprule
Scene type & Efficient-UNet & GeoUQ-GFNet & Nearest & ResNet-UNet & ViT-UNet \\
\midrule
Building-medium & 5.6122 & \textbf{5.3739} & 7.8389 & 5.7396 & 7.1343 \\
Building-sparse & 6.0882 & \textbf{5.7654} & 9.0199 & 6.1814 & 6.8790 \\
Canyon & 5.7237 & \textbf{5.4255} & 8.0844 & 5.8100 & 6.9207 \\
Crossroad & 5.4449 & \textbf{5.2028} & 7.5684 & 5.5203 & 6.9970 \\
Offset-crossroad & 4.9781 & \textbf{4.7831} & 7.0034 & 5.0875 & 6.6338 \\
T-junction & 5.5080 & \textbf{5.3109} & 7.8881 & 5.6255 & 7.0008 \\
\midrule
Average over scenes & 5.5592 & \textbf{5.3102} & 7.9005 & 5.6607 & 6.9276 \\
\bottomrule
\end{tabular}
\end{table*}

\subsection{Performance Evaluation Metrics}
\label{subsec_metrics}

We evaluate reconstruction quality using both normalized-domain and physical-domain metrics. Let $\hat{\mathbf{G}}$ and $\mathbf{G}$ denote the predicted and ground-truth gain maps, respectively. All metrics are computed over the accessible region, and for training-related analysis the error on unobserved accessible cells is emphasized.

The normalized root mean square error RMSE is defined as
\begin{equation}
\mathrm{RMSE}_{\mathrm{norm}}
=
\sqrt{
\frac{
\sum_{i,j} M_{a,i,j} \left( \hat{G}^{\mathrm{norm}}_{i,j} - G^{\mathrm{norm}}_{i,j} \right)^2
}{
\sum_{i,j} M_{a,i,j}
}
}.
\end{equation}

In the physical domain, we report the RMSE and Mean Absolute Error (MAE) in dB:
\begin{equation}
\mathrm{RMSE}_{\mathrm{dB}}
=
\sqrt{
\frac{
\sum_{i,j} M_{a,i,j} \left( \hat{G}_{i,j} - G_{i,j} \right)^2
}{
\sum_{i,j} M_{a,i,j}
}
},
\end{equation}
\begin{equation}
\mathrm{MAE}_{\mathrm{dB}}
=
\frac{
\sum_{i,j} M_{a,i,j} \left| \hat{G}_{i,j} - G_{i,j} \right|
}{
\sum_{i,j} M_{a,i,j}
}.
\end{equation}

For uncertainty-aware models, we additionally evaluate the usefulness of the predicted uncertainty map through the correlation between absolute prediction error and predicted standard deviation. A higher positive correlation indicates that the model assigns larger uncertainty to regions that are indeed harder to reconstruct, which is important for active sensing \cite{Lee2019AdaptiveBayesianRadioTomography,Polyzos2024BayesianActiveLearningRadioMap}.

For active measurement selection, we report the reconstruction RMSE after each additional measurement budget increment. This metric directly reflects the practical value of uncertainty-guided querying under a fixed sensing budget. In the comparison, uncertainty-guided selection and random selection are evaluated under exactly the same incremental budget, which makes the resulting performance gap a direct indicator of decision utility \cite{Polyzos2024BayesianActiveLearningRadioMap,Shrestha2023SpectrumSurveying}.

\section{Results and Analysis}
\label{sec_results}

This section evaluates scene level generalization, overall reconstruction accuracy, BS level robustness, and the practical value of uncertainty guided active measurement selection. Unless otherwise stated, all numerical comparisons in the text follow the results summarized in Tables~\ref{tab:scene_level_rmse}, \ref{tab:overall_main_results}, and \ref{tab:active_final_results}.

\begin{figure}[htbp]
    \centering
    \includegraphics[width=\columnwidth]{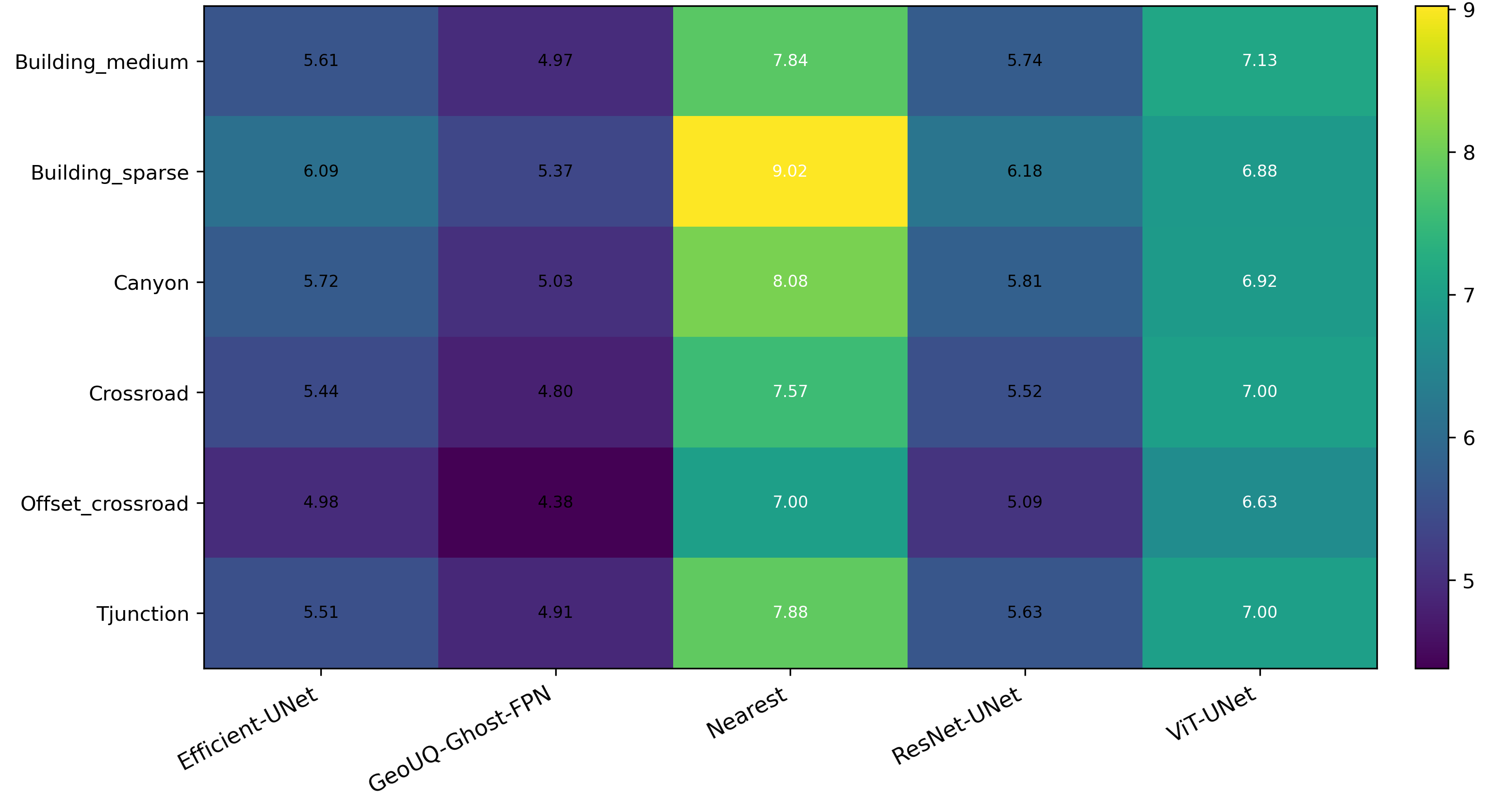}
    \caption{Scene wise RMSE heatmap in dB.}
    \label{fig:scene_rmse_heatmap}
\end{figure}

\begin{figure}[htbp]
    \centering
    \includegraphics[width=\columnwidth]{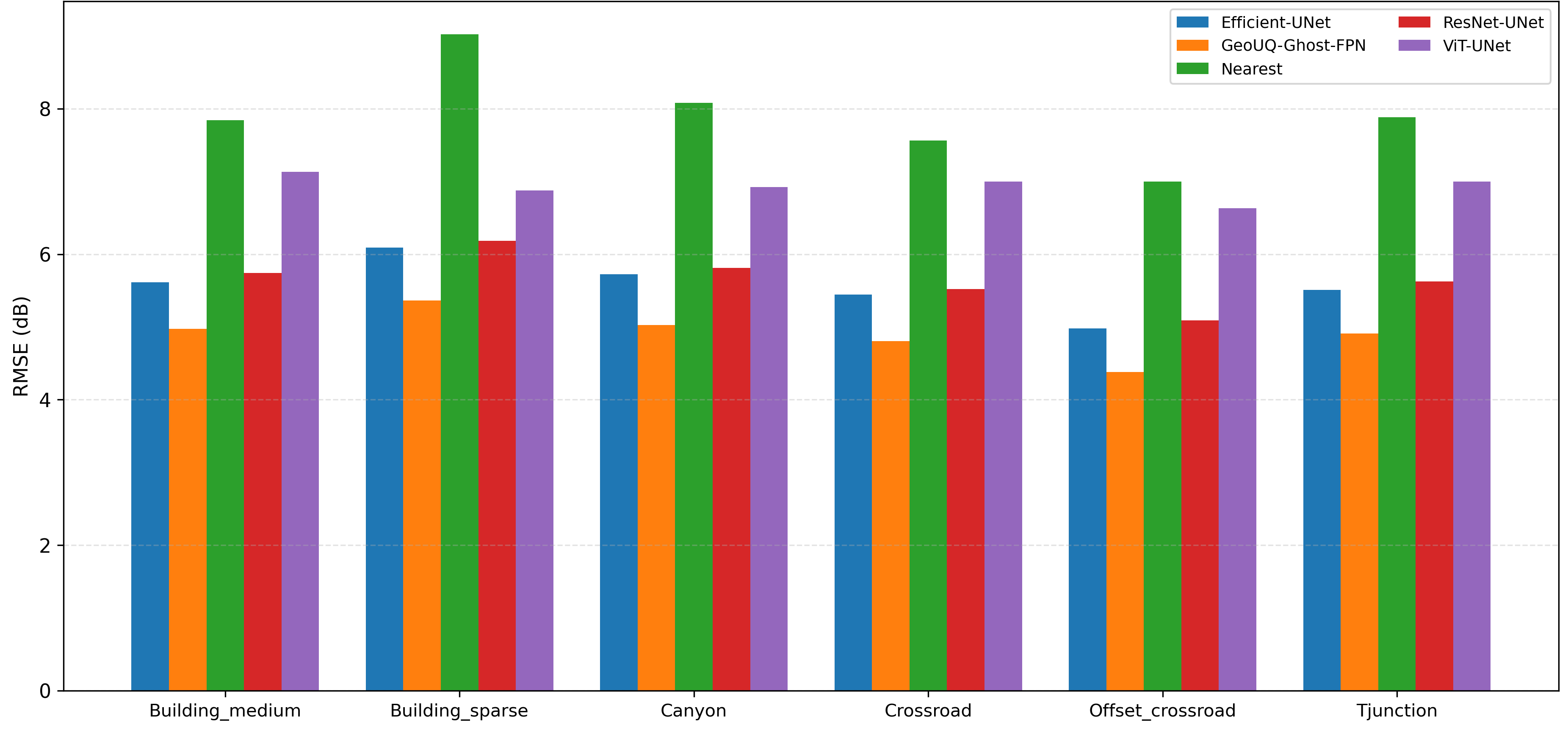}
    \caption{RMSE grouped by scene type.}
    \label{fig:scene_group_bar}
\end{figure}

The scene dependent behavior is first reported in Table~\ref{tab:scene_level_rmse}, Fig.~\ref{fig:scene_rmse_heatmap}, and Fig.~\ref{fig:scene_group_bar}. Several clear patterns can be observed. First, GeoUQ-GFNet remains the best model in every scene category. Its average RMSE over all scene types is $5.3102$ dB, which is lower than Efficient-UNet at $5.5592$ dB, ResNet-UNet at $5.6607$ dB, ViT-UNet at $6.9276$ dB, and Nearest at $7.9005$ dB. Second, scene difficulty is not uniform. Building-sparse is the hardest family for all methods, while Offset-crossroad is the easiest. For example, GeoUQ-GFNet obtains $5.7654$ dB in Building-sparse but only $4.7831$ dB in Offset-crossroad. The same trend appears for all other models. This is physically consistent with the geometry of the scenes. Sparse building layouts contain both broad open regions and sharp blockage transitions, which makes the gain field harder to infer from limited observations. In contrast, the Offset-crossroad layout has a more regular geometric structure, so the dominant spatial propagation trend is easier to recover. Third, the nearest interpolation baseline is consistently much weaker than the learned models in every scene. This confirms that sparse radio map reconstruction should not be viewed as a purely local interpolation problem. Since the missing area is shaped by geometry, visibility, and observation support, our results highlight that learning propagation aware spatial structure is necessary.

\begin{figure*}[!t]
    \centering
    \begin{subfigure}[t]{0.32\textwidth}
        \centering
        \includegraphics[height=4.8cm, keepaspectratio]{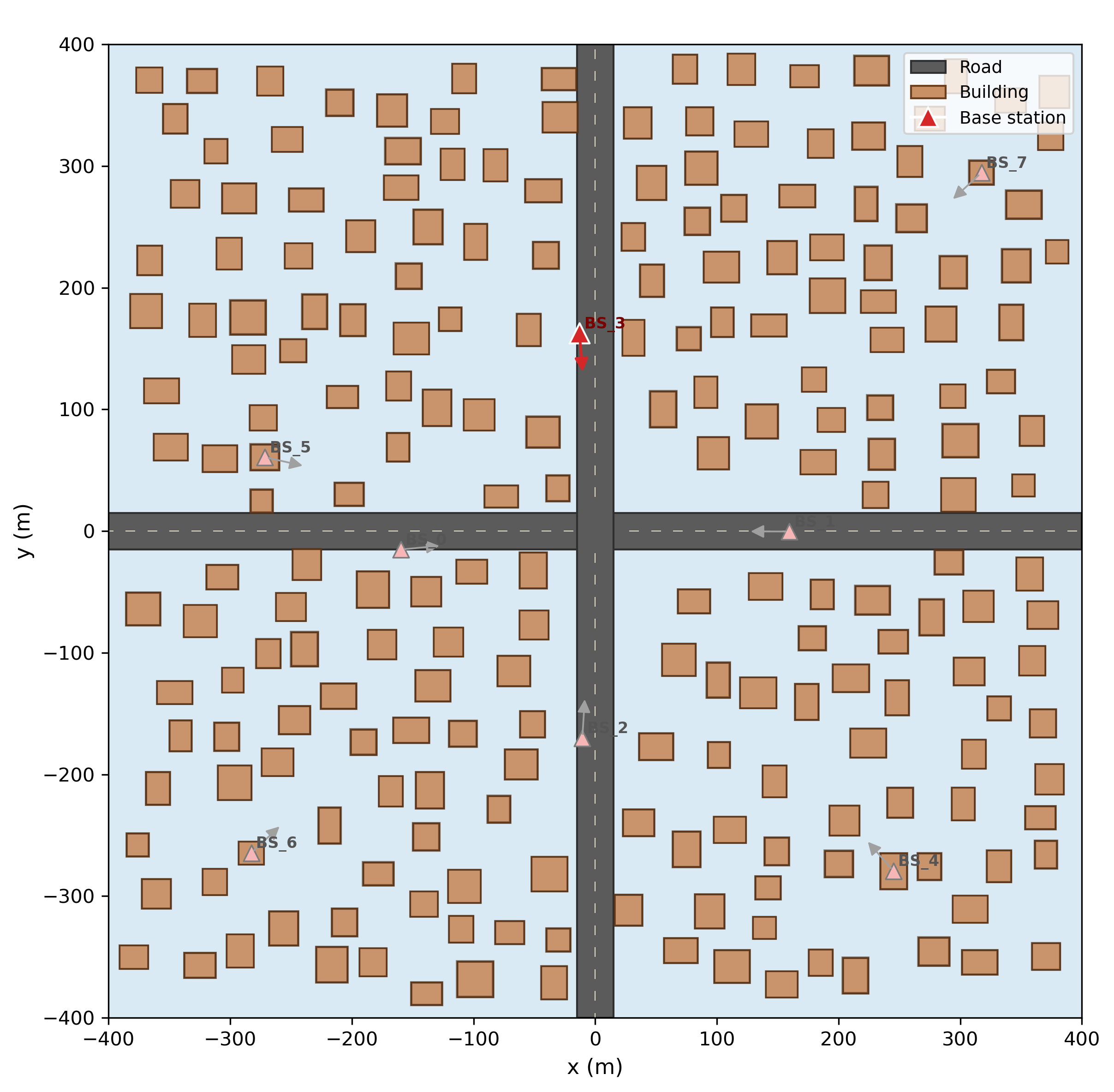}
        \caption{Scene layout and BS placement.}
        \label{fig:qual_layout}
    \end{subfigure}
    \hfill
    \begin{subfigure}[t]{0.32\textwidth}
        \centering
        \includegraphics[height=4.8cm, keepaspectratio]{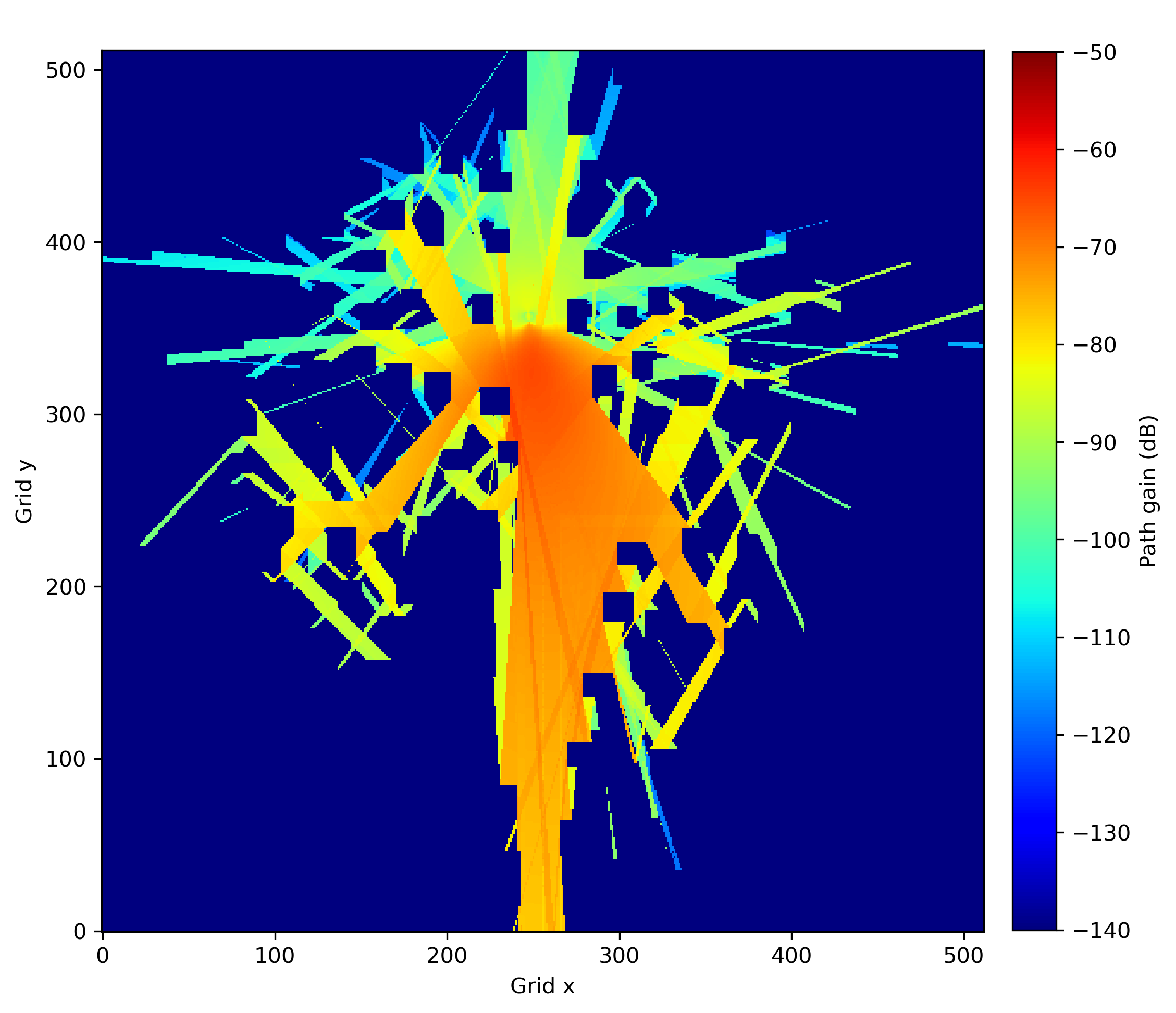}
        \caption{Ground truth radio map.}
        \label{fig:qual_gt}
    \end{subfigure}
    \hfill
    \begin{subfigure}[t]{0.32\textwidth}
        \centering
        \includegraphics[height=4.8cm, keepaspectratio]{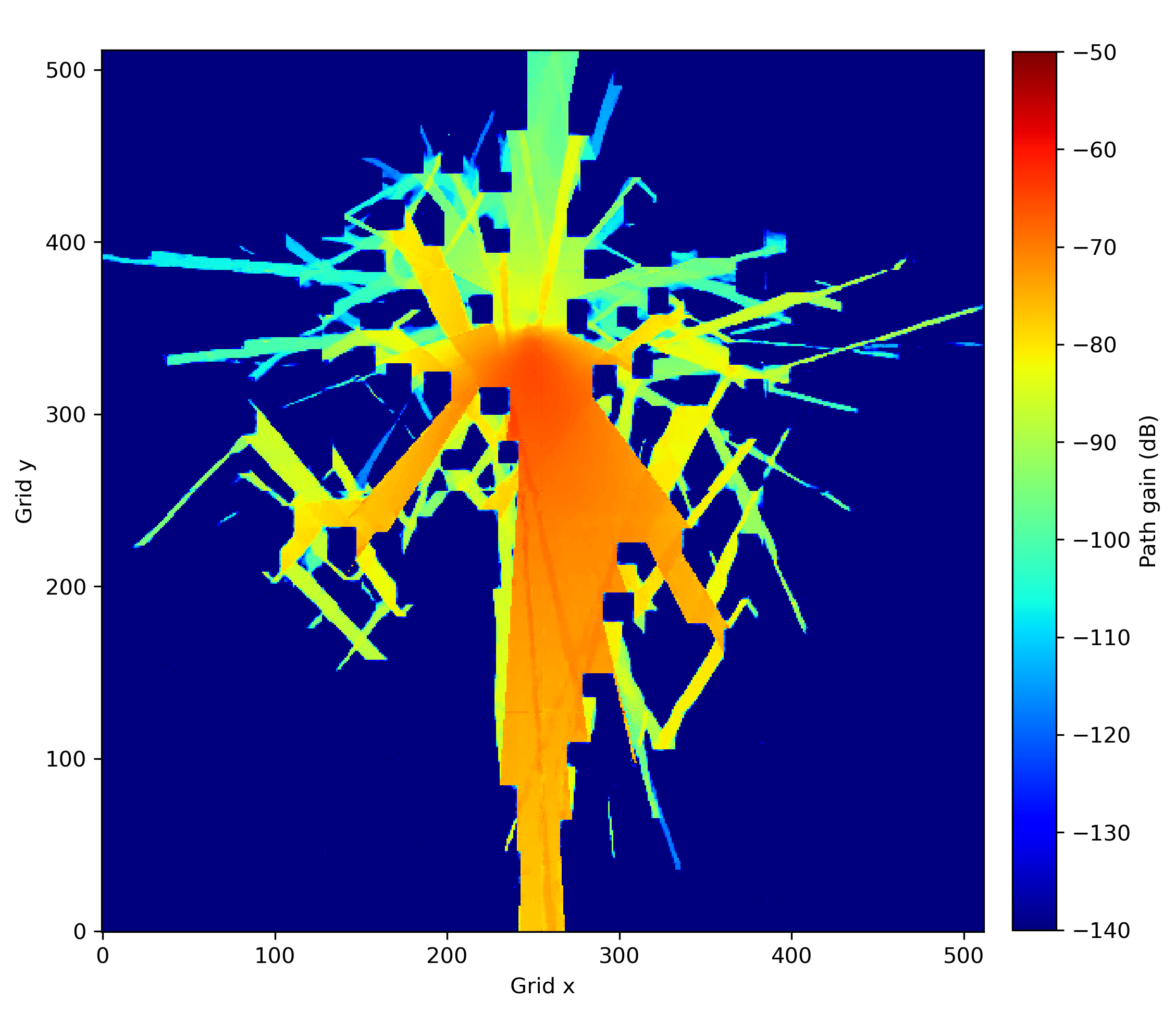}
        \caption{GeoUQ-GFNet prediction.}
        \label{fig:qual_pred}
    \end{subfigure}
    \caption{Representative qualitative example for a crossroad scene. The proposed model closely recovers the dominant propagation corridor and the main angular energy structure of the ground truth radio map.}
    \label{fig:qualitative_case}
\end{figure*}

\begin{table}[htbp]
\centering
\caption{Overall Reconstruction Performance}
\label{tab:overall_main_results}
\setlength{\tabcolsep}{4pt}
\begin{tabular}{lccc}
\toprule
Model & RMSE (dB) & MAE (dB) & Error uncertainty corr. \\
\midrule
GeoUQ-GFNet & \textbf{7.1025} & \textbf{1.6191} & \textbf{0.5818} \\
Efficient-UNet & 7.7557 & 2.0449 & 0.5780 \\
ViT-UNet & 7.7895 & 2.0978 & 0.5814 \\
ResNet-UNet & 7.8529 & 2.1139 & 0.5813 \\
Nearest & 11.9412 & 3.7052 & 0.0000 \\
\bottomrule
\end{tabular}
\end{table}

The overall comparison is then summarized in Table~\ref{tab:overall_main_results}. GeoUQ-GFNet achieves the best RMSE of $7.1025$ dB, the best MAE of $1.6191$ dB, and the best error uncertainty correlation of $0.5818$. Efficient-UNet is the second strongest model, with RMSE $7.7557$ dB and MAE $2.0449$ dB. ViT-UNet and ResNet-UNet follow closely, and both remain clearly better than the nearest interpolation baseline. The improvement of GeoUQ-GFNet is substantial rather than marginal. Relative to Efficient-UNet, the RMSE is reduced by about $0.65$ dB. Relative to ViT-UNet and ResNet-UNet, the reduction is about $0.69$ dB and $0.75$ dB, respectively. The MAE ranking is fully consistent with the RMSE ranking. These results show that the proposed design improves both global reconstruction fidelity and local error magnitude. It is also important that the uncertainty branch does not weaken reconstruction quality. Instead, the same model attains the best reconstruction metrics and the highest error uncertainty correlation, which supports the core claim of the paper that uncertainty estimation can be introduced without sacrificing one shot prediction quality.

\begin{figure}[htbp]
    \centering
    \includegraphics[width=\columnwidth]{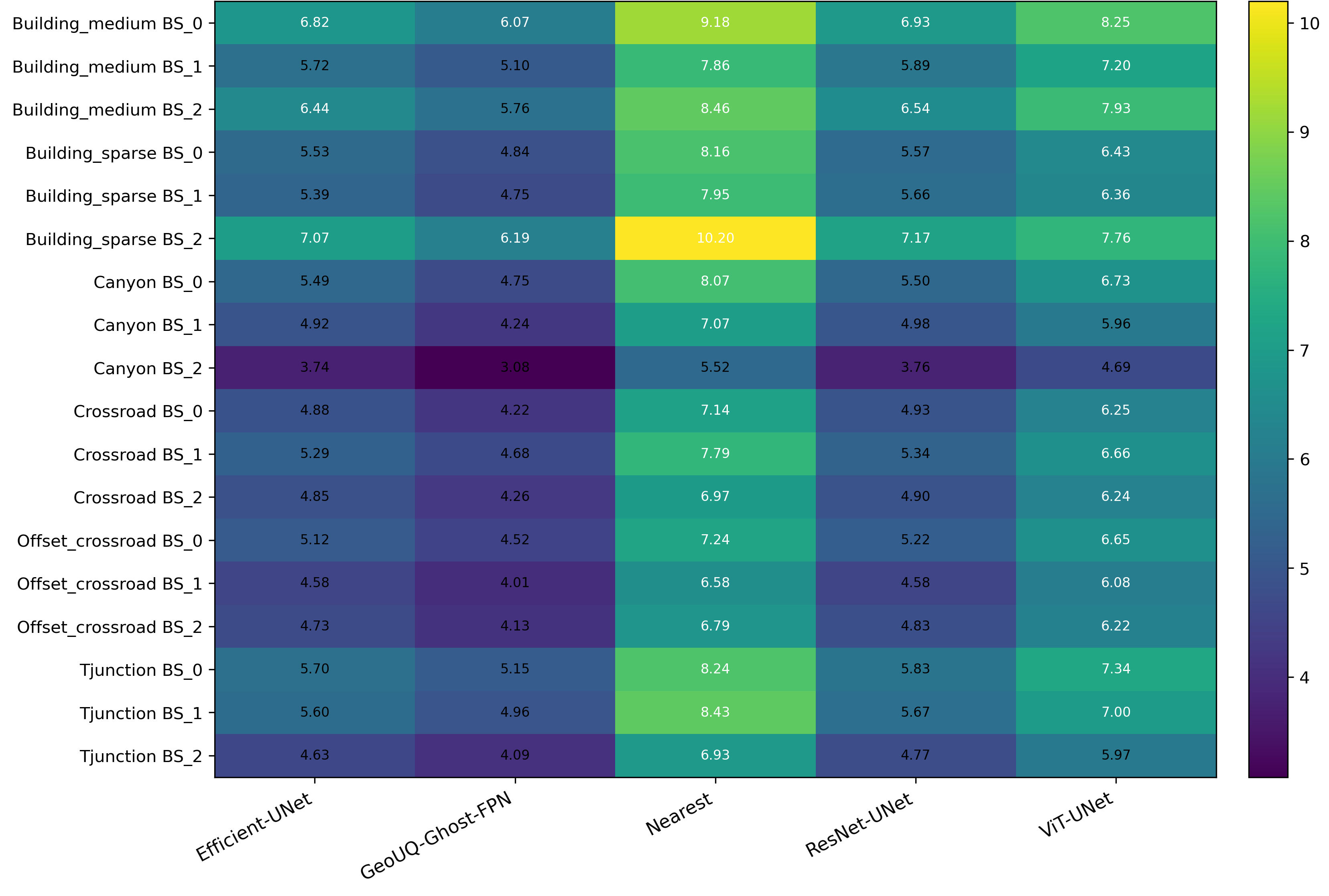}
    \caption{BS wise RMSE heatmap in dB in each scene.}
    \label{fig:bs_heatmap_top3}
\end{figure}

Beyond the average metrics, it is also useful to inspect a representative qualitative example. Fig.~\ref{fig:qualitative_case} shows one crossroad scene together with its ground truth radio map and the corresponding prediction of GeoUQ-GFNet. Several observations can be made. First, the predicted map preserves the dominant propagation structure radiating from the selected BS and accurately captures the strong corridor aligned with the road geometry. Second, the major energy branches and blockage induced angular transitions are well maintained, indicating that the model does not merely smooth the field but is able to recover physically meaningful spatial structure. Third, although minor local discrepancies remain in some weak reflection regions and fine branches, the overall topology of the gain field is highly consistent with the ground truth. This qualitative case complements the quantitative results above and visually demonstrates that the proposed model can reconstruct not only average intensity levels but also the main geometric propagation pattern of a realistic scene.

A finer analysis is provided by the BS level heatmap in Fig.~\ref{fig:bs_heatmap_top3}. Here we keep the representative top three BS instances from each scene, which makes the comparison much easier to read than the full BS list while still preserving the cross scene trend. The main conclusion remains unchanged. GeoUQ-GFNet gives the lowest RMSE for nearly all listed BS cases, and Nearest remains the weakest baseline throughout. This result is important because it shows that the gain of the proposed model is not concentrated on only a few favorable transmitters. Instead, the advantage persists across different BS placements and across both road oriented and building dominated environments. The BS wise view also reveals meaningful intra scene variation. In the Canyon family, BS\_2 is much easier than BS\_0 and BS\_1 for every model. In the Building-sparse family, BS\_2 is distinctly more difficult than BS\_0 and BS\_1. This is expected because changing the BS location changes the angular visibility, the blockage configuration, and the balance between line of sight and non line of sight regions. Therefore, scene averages alone are not sufficient to characterize sparse radio map difficulty. A strong model must remain robust to both scene topology and BS placement, and the results in Fig.~\ref{fig:bs_heatmap_top3} support that conclusion for GeoUQ-GFNet.

\begin{table}[htbp]
\centering
\caption{Active sensing comparison at budget ratio $0.04$}
\label{tab:active_final_results}
\setlength{\tabcolsep}{4pt}
\begin{tabular}{lccc}
\toprule
Model & UQ RMSE (dB) & Rand. RMSE (dB) & Gain (dB) \\
\midrule
GeoUQ-GFNet & \textbf{3.1002} & \textbf{5.9144} & \textbf{2.8142} \\
ViT-UNet & 3.4558 & 6.2599 & 2.8041 \\
Efficient-UNet & 3.5101 & 6.2742 & 2.7641 \\
ResNet-UNet & 3.5543 & 6.3106 & 2.7563 \\
Nearest & 9.8791 & 9.8791 & 0.0000 \\
\bottomrule
\end{tabular}
\end{table}

\begin{figure}[htbp]
    \centering
    \includegraphics[width=\columnwidth]{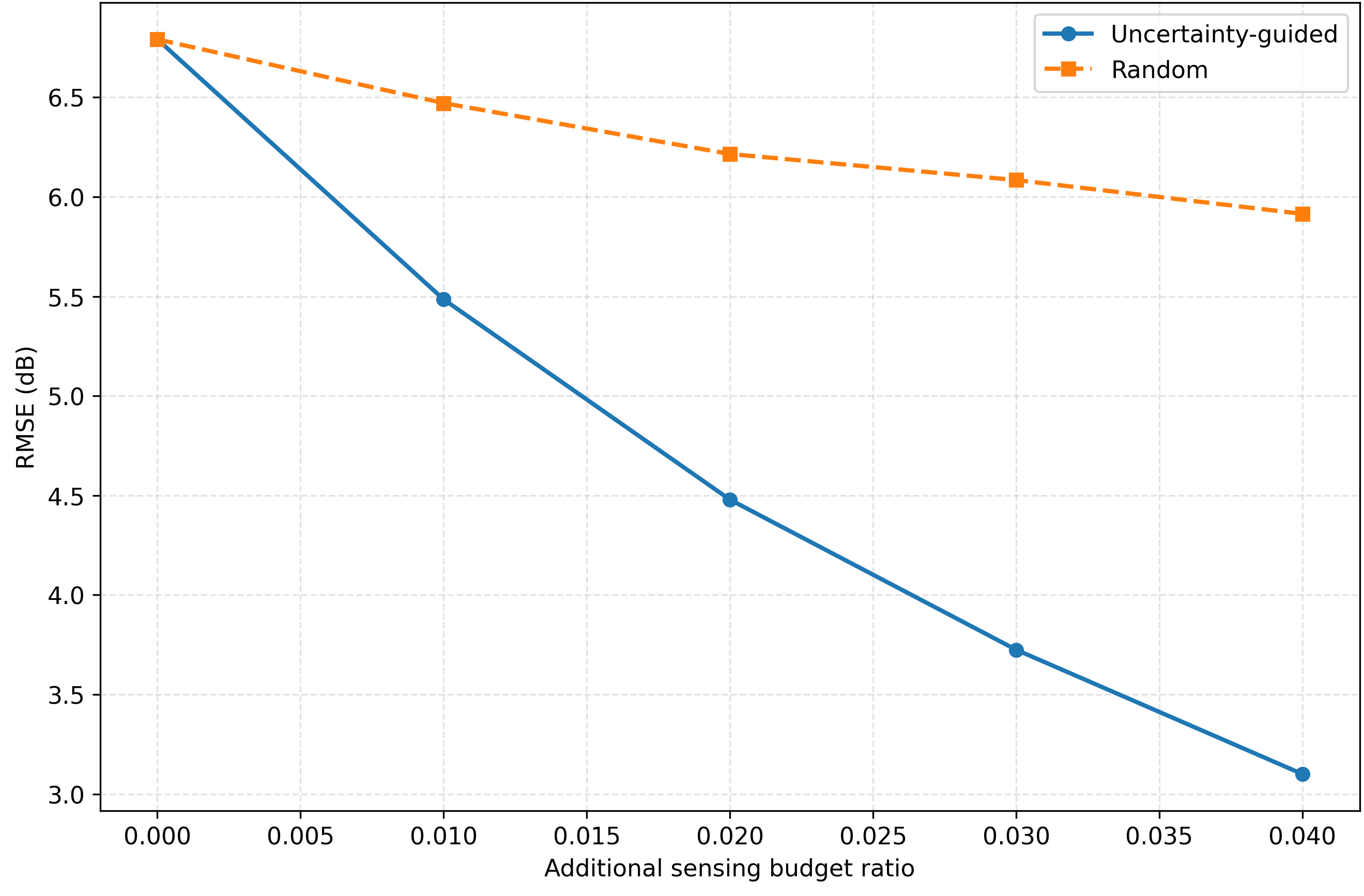}
    \caption{Active sensing result of GeoUQ-GFNet. Uncertainty guided querying steadily reduces RMSE as the additional sensing budget increases and clearly outperforms random selection.}
    \label{fig:active_geouq}
\end{figure}

The most practically important result concerns active measurement selection. The final comparison is summarized in Table~\ref{tab:active_final_results}, and the full budget trajectory of GeoUQ-GFNet is shown in Fig.~\ref{fig:active_geouq}. Several observations are especially important. First, uncertainty guided sensing produces a large gain for all learned models. At the final additional sensing budget ratio of $0.04$, GeoUQ-GFNet reaches $3.1002$ dB, while random selection remains at $5.9144$ dB. This corresponds to a gain of $2.8142$ dB. Efficient-UNet, ResNet-UNet, and ViT-UNet show similarly large gains of $2.7641$ dB, $2.7563$ dB, and $2.8041$ dB, respectively. Second, the nearest baseline shows no gain because it does not provide a useful uncertainty estimate and therefore cannot support informed query selection. Third, the improvement curve of GeoUQ-GFNet is smooth and monotonic over the full sensing process. The RMSE decreases from $6.79$ dB to $5.49$ dB, then to $4.48$ dB, then to $3.73$ dB, and finally to $3.10$ dB. In contrast, the random policy only decreases from $6.79$ dB to $5.91$ dB over the same budget range. This result directly supports the central argument of the paper. The main value of predictive uncertainty lies not only in confidence reporting, but more importantly in guiding where to measure next under a limited sensing budget. In other words, uncertainty has strong decision utility, and this utility is larger than the gap seen in one shot reconstruction metrics alone.

Taken together, the results support the following three novel conclusions for sparse gain radio map reconstruction: (1) GeoUQ-GFNet is the strongest overall model in reconstruction accuracy and uncertainty quality. (2) Scene geometry and BS placement both have a strong effect on reconstruction difficulty, so evaluation should be performed beyond a single average number. (3) Most importantly, uncertainty is especially valuable as a decision signal for measurement acquisition. This is why sparse radio map reconstruction should be studied not only as a spatial prediction task, but also as a measurement aware decision problem.

\section{Conclusion}
\label{sec_conclusion}

This paper studied sparse gain radio map reconstruction under geometry and sensing constraints. We constructed a controllable ray tracing benchmark, termed UrbanRT-RM, with multiple scenes, BS deployments, and sampling modes, and developed GeoUQ-GFNet for joint gain reconstruction and uncertainty estimation. The results show that GeoUQ-GFNet achieves the best overall reconstruction performance and remains robust across different scene types and BS placements. They also show that reconstruction difficulty depends strongly on scene geometry and transmitter location, and that predictive uncertainty is especially useful for active measurement selection, where uncertainty guided querying yields much larger gains than one shot reconstruction improvement alone. These findings suggest that sparse radio map reconstruction should be treated not only as a spatial prediction problem but also as a geometry aware and measurement aware decision problem.



\end{document}